\documentclass[a4paper,fleqn]{cas-dc}

\usepackage[numbers,sort&compress]{natbib}
\usepackage{enumitem}
\usepackage{framed}
\usepackage{algorithm}
\usepackage{algpseudocode}
\usepackage{graphicx}
\hypersetup{hypertexnames=false}
\usepackage{colortbl}
\newcommand{\bestresult}[1]{\cellcolor{gray!15}\textbf{#1}}

\begin{document}
\let\WriteBookmarks\relax
\def\floatpagepagefraction{1}
\def\textpagefraction{.001}

\shorttitle{EPPC-OASIS with structured inference}
\shortauthors{Fodeh et al.}

\title[mode=title]{EPPC-OASIS: Ontology-Aware Adaptation and Structured Inference Refinement for Electronic Patient-Provider Communication Mining in Secure Messages}

%\title[mode=title]{EPPC-OASIS: Ontology-Aware Large Language Models for Mining Electronic Patient-Provider Communication in Secure Messages}

%\title[mode=title]{EPPC-OASIS: Ontology-Aware Large Language Models for Structured Extraction of Electronic Patient-Provider Communication}

% EPPC-OASIS: Ontology-Aware Structured Information Synthesis for Electronic Patient-Provider Communication Mining

\author[1]{Samah Fodeh}
\cormark[1]
\ead{samah.fodeh@yale.edu}

\author[1]{Sreeraj Ramachandran}
\author[1]{Elyas Irankhah}
\author[1]{Muhammad Arif}
\author[1]{Afshan Khan}
\author[1]{Ganesh Puthiaraju}
\author[1]{Linhai Ma}
\author[1]{Srivani Talakokkul}
\author[2]{Jordan Alpert}
\author[3]{Sarah Schellhorn}

\affiliation[1]{
  organization={Yale University},
  city={New Haven},
  state={CT},
  country={USA}
}

\affiliation[2]{
  organization={Cleveland Clinic Lerner College of Medicine of Case Western Reserve University, Cleveland Clinic},
  city={Cleveland},
  state={OH},
  country={USA}
}

\affiliation[3]{
  organization={Medical Oncology, Yale School of Medicine},
  city={New Haven},
  state={CT},
  country={USA}
}

\cortext[1]{Corresponding author: Samah Fodeh, Yale University, New Haven, CT, USA.}

\begin{abstract}
Secure patient-provider messages contain clinically important communication behaviors that are difficult to characterize manually at scale. The Electronic Patient-Provider Communication (EPPC) framework provides an ontology for coding these behaviors, but automated extraction remains challenging because predictions must preserve fine-grained code/sub-code structure while grounding annotations in message text. We developed EPPC-OASIS, an ontology-aware adaptation approach for structured EPPC extraction, and combined it with deployable inference-refinement procedures designed to improve the coherence of final annotations. EPPC-OASIS augments supervised fine-tuning with a Wasserstein alignment objective that encourages alignment between model representation neighborhoods and EPPC ontology-derived neighborhoods, while inference refinement uses verification, self-consistency, hybrid correction, and selection or ensembling to address residual prediction errors. %EPPC-OASIS augments supervised fine-tuning with a Wasserstein alignment objective that matches model representation neighborhoods to EPPC ontology-derived neighborhoods, while inference refinement uses verification, self-consistency, hybrid correction, and selection or ensembling to address residual prediction errors. 
We evaluated the framework on a de-identified corpus of secure patient-provider messages against prompting, supervised fine-tuning, preference-based, and robustness-oriented baselines across multiple open-weight language models. Across model families, the best deployable pipeline achieved \textbf{77.13\%} Code+Sub-code F1 and \textbf{63.83\%} Triplet F1, corresponding to modest but consistent absolute gains of \textbf{+1.39} and \textbf{+2.12} F1 points over the strongest supervised fine-tuning baseline. %The best deployable pipeline achieved \textbf{77.13\%} Code+Sub-code F1 and \textbf{63.83\%} Triplet F1, corresponding to absolute gains of \textbf{+1.39} and \textbf{+2.12} F1 points over the strongest supervised fine-tuning baseline.
These results suggest that ontology-aware adaptation with structured inference refinement can support scalable retrospective EPPC mining, although external validation is needed before operational use.
\end{abstract}

% \begin{highlights}
% \item EPPC-OASIS aligns model representations with ontology neighborhoods.
% \item Inference refinement combines self-consistency and label correction.
% \item Ontology-aware adaptation improves hierarchical EPPC extraction.
% \item Ontology-aware adaptation enables structured clinical message mining.
% \end{highlights}


\maketitle

\section{Introduction}
\label{sec:introduction}

Secure patient-provider messaging has become a routine part of outpatient care, creating a longitudinal record of clinical communication outside the traditional visit encounter\cite{cronin2015securemessaging, huang2022securemessages}. These messages contain not only requests for information and clinician responses, but also evidence of care coordination, logistical barriers, emotional concerns, social needs, and shared decision-making\cite{north2020securemessages}. The Electronic Patient-Provider Communication (EPPC) framework provides a structured way to characterize these communication behaviors \cite{fodeh2026eppcminerben, fodeh2026pvminer, fodeh2026pvminerllm, fodeh2026tab, fodeh2026stardro}, but manual EPPC coding is difficult to scale across large message corpora. Reliable automated EPPC extraction could therefore support retrospective communication research, quality measurement, and cohort-level analysis of how patients and care teams exchange information over time\cite{agrawal2022clinicalie,wang2022hpt,u2023instances,welleck2022selfcorrect}.

Automated EPPC extraction is challenging because the task requires more than assigning a single label to a message \cite{tsoumakas2007multilabel}. Each prediction must identify the communication behaviors expressed in the text, map them to the appropriate high-level EPPC codes and sub-codes, and provide supporting evidence from the original message\cite{deyoung2020eraser,agrawal2022clinicalie}. These requirements are difficult to satisfy jointly. EPPC labels are hierarchical, imbalanced, and often semantically adjacent, while individual message segments may contain multiple communication behaviors expressed in compact or informal language\cite{xu2021hierarchical}. Consequently, model errors are often structured rather than random: a prediction may recover the general communicative intent but choose the wrong sub-code, miss a low-frequency behavior, or assign a correct label while grounding it in incomplete or mismatched evidence\cite{henning2023classimbalance,blanchard2022keyword}.

Recent work has established EPPC extraction as a benchmark task for evaluating large language models on structured patient-provider communication coding, showing that modern instruction-tuned models can recover clinically meaningful communication behaviors but still struggle with fine-grained labels and evidence grounding \cite{fodeh2026eppcminerben}. Related robustness-oriented work has further shown that performance is uneven across difficult or underrepresented portions of the EPPC label space \cite{fodeh2026stardro}. Together, these studies motivate the use of LLMs for scalable EPPC mining, while also exposing two methodological gaps. First, standard prompting or supervised fine-tuning treats each target annotation largely as an output sequence, rather than using the EPPC ontology to shape how related examples and labels should be represented during adaptation. Second, inference is commonly treated as a single generation step, even though EPPC errors often separate into label-consistency failures and evidence-grounding failures that may benefit from different correction mechanisms.

In this study, we develop and evaluate a two-stage approach for structured EPPC extraction from de-identified patient-provider messages. The first stage, EPPC-OASIS, augments supervised fine-tuning with ontology-aware Wasserstein alignment so that the model is trained not only to reproduce target annotations, but also to align learned representation neighborhoods with neighborhoods implied by the EPPC code and sub-code inventory \cite{cuturi2013sinkhorn}. The second stage applies deployable structured inference refinement, using verification, self-consistency, hybrid correction, and selection procedures to address residual errors in label assignment and evidence grounding \cite{wei2023cot,wang2023selfconsistency,madaan2023selfrefine}. We evaluate the approach using deployable Code/Sub-code and triplet-level extraction metrics, with the goal of improving ontology-consistent EPPC mining for retrospective clinical communication research.

This study makes four contributions:
\begin{enumerate}[leftmargin=*]
\item We introduce EPPC-OASIS, an ontology-aware adaptation method that uses the structure of the EPPC code and sub-code inventory to define a Wasserstein alignment objective during fine-tuning.
\item We develop a structured inference-refinement framework that combines self-verification, self-consistency, hybrid label correction, and deployable selection or ensembling to improve the final structured EPPC annotation set.
\item We evaluate the resulting pipelines against prompting, supervised fine-tuning, preference-based, and robustness-oriented baselines across multiple open-weight model families.
\item We provide ablations and diagnostics that separate the effects of ontology-aware training from inference-time refinement and distinguish deployable pipeline performance from component-wise diagnostic upper bounds.
\end{enumerate}

\section{Methods}
\label{sec:methods}

\subsection{Study Design and Setting}
\label{subsec:study-design}

\begin{figure*}[t]
\centering
\includegraphics[width=\textwidth]{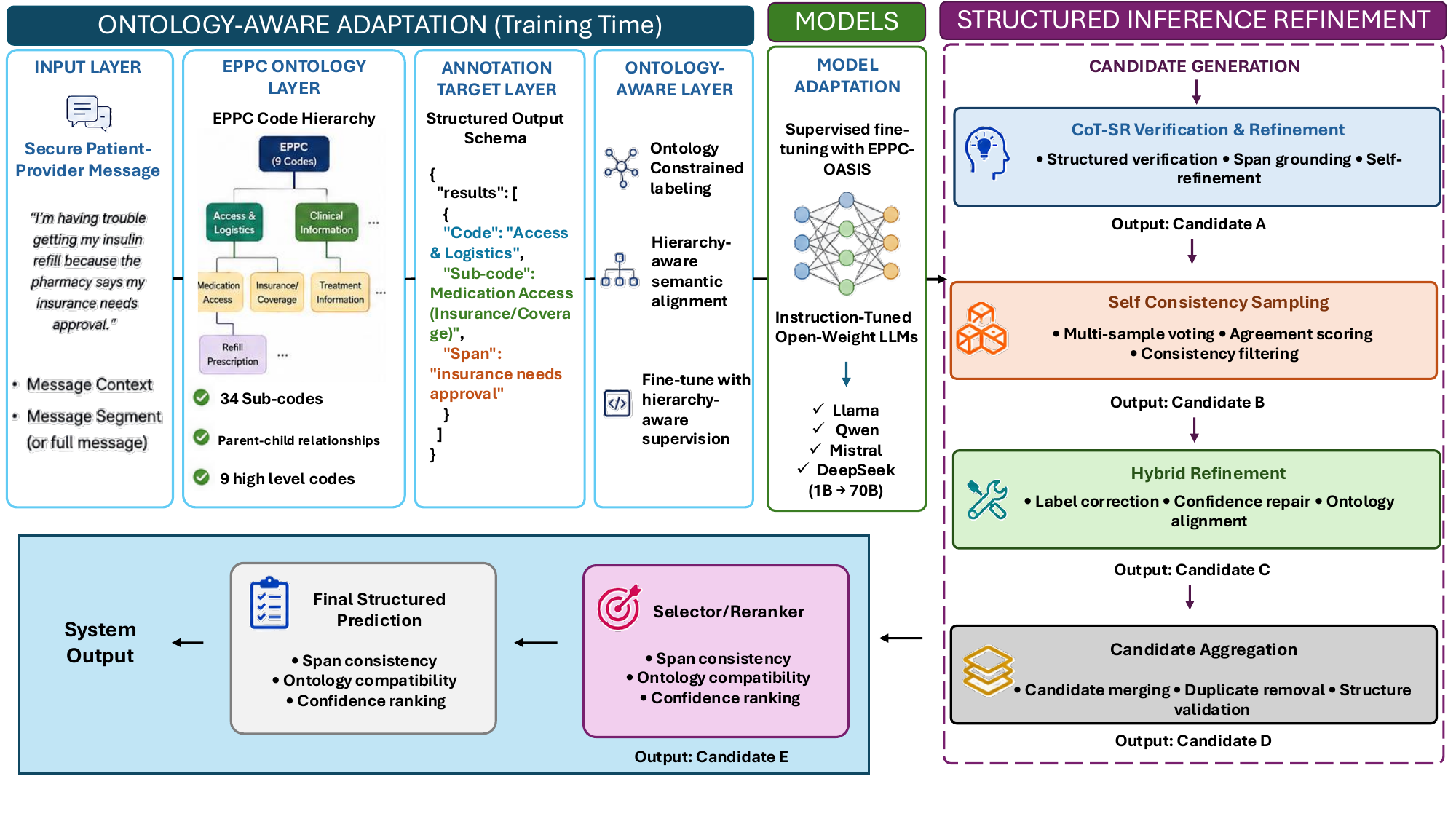}
\caption{
Overview of the proposed EPPC-OASIS framework for structured EPPC extraction from secure patient-provider messages. The workflow consists of two major stages: ontology-aware adaptation during training and structured inference refinement during inference. The training stage converts secure-message inputs into structured JSON annotations using the EPPC ontology hierarchy and ontology-aware representation alignment. Instruction-tuned open-weight LLMs are then adapted using supervised fine-tuning with ontology-constrained learning signals. During inference, the framework applies multiple structured refinement strategies, including Chain of Thought-Self Refinement(CoT-SR) verification and refinement, self-consistency sampling, hybrid refinement, candidate aggregation, and selector/reranker-based prediction selection. The final output is a schema-valid structured EPPC prediction containing Code, Sub-code, and supporting evidence Span annotations.
}

\label{fig:eppc_oasis_workflow}
\end{figure*}

This study was designed as a retrospective methodological evaluation of automated EPPC extraction from de-identified secure patient-provider messages. Within the EPPC framework, each annotated message segment was treated as a structured communication-coding instance, allowing us to evaluate how well language-model methods could recover ontology-governed communication behaviors from previously coded data. All model development used fixed training data, and final performance was assessed on a held-out test split that was not used for model selection. The study therefore focused on method development and retrospective informatics use, rather than clinical decision support deployment or generation of patient-facing recommendations. Figure~\ref{fig:eppc_oasis_workflow} presents the overall study workflow, including ontology-aware adaptation, structured inference refinement.

\subsection{EPPC Annotation Schema and Prediction Target}
\label{subsec:task-schema}

The present study builds on the EPPC annotation framework by treating each secure-message excerpt as a structured clinical communication coding task. For each instance, the model was given the relevant patient-provider message context and the sentence or message segment to be coded. The expected output was a JSON annotation listing any EPPC categories expressed in that segment. Each annotation included a high-level \texttt{Code}, a more specific \texttt{Sub-code}, and a supporting text \texttt{Span} copied from the original message. The \texttt{Code} and \texttt{Sub-code} identify the communication behavior, while the \texttt{Span} grounds that label in the source text. This formulation preserves the structure of manual EPPC coding while allowing the task to be evaluated as automated structured extraction.

To make model outputs comparable with manual annotations, all predictions were evaluated against a common structured schema. A valid response was defined as a JSON object with a \texttt{results} field containing a list of annotations, where each annotation included a \texttt{Code}, \texttt{Sub-code}, and \texttt{Span}. The \texttt{Code} and \texttt{Sub-code} were required to come from the predefined EPPC label inventory, with each sub-code interpreted in relation to its parent code. Segments with no applicable EPPC category were represented by an empty \texttt{results} list.

Although the task has a structured output format, we did not use constrained decoding in the primary experiments. Instead, models generated responses normally, and the resulting text was parsed into the target schema before scoring. This choice was based on preliminary comparisons in which guided decoding improved JSON well-formedness but did not improve the extraction metrics used in this study. We therefore used lightweight post hoc recovery for minor formatting deviations, including extracting JSON from Markdown code fences or from surrounding explanatory text. Outputs that could not be parsed after this recovery step were treated as invalid predictions.

%\placeholderfigure{Overview of the EPPC structured extraction task. The final figure should show a secure-message context, the current sentence or extraction unit, and the target JSON output with Code, Sub-code, and supporting Span fields.}{fig:task-overview}{Placeholder for Figure 1: EPPC task overview}

\subsection{Dataset}
\label{subsec:dataset}

The dataset used in this study was derived from de-identified secure patient-provider messages from the Yale New Haven Hospital patient portal, following the EPPC data source and annotation framework established in prior EPPCMinerBen work \cite{fodeh2026eppcminerben}. The corpus captures asynchronous communication between patients and clinical teams, including informational exchange, care coordination, socio-emotional communication, partnership-building, and shared decision-making behaviors. Each annotated instance was converted into a structured extraction example, with the message context provided as input and the corresponding EPPC annotations provided as the target output.

\begin{figure*}[t]
\centering
\includegraphics[width=\textwidth]{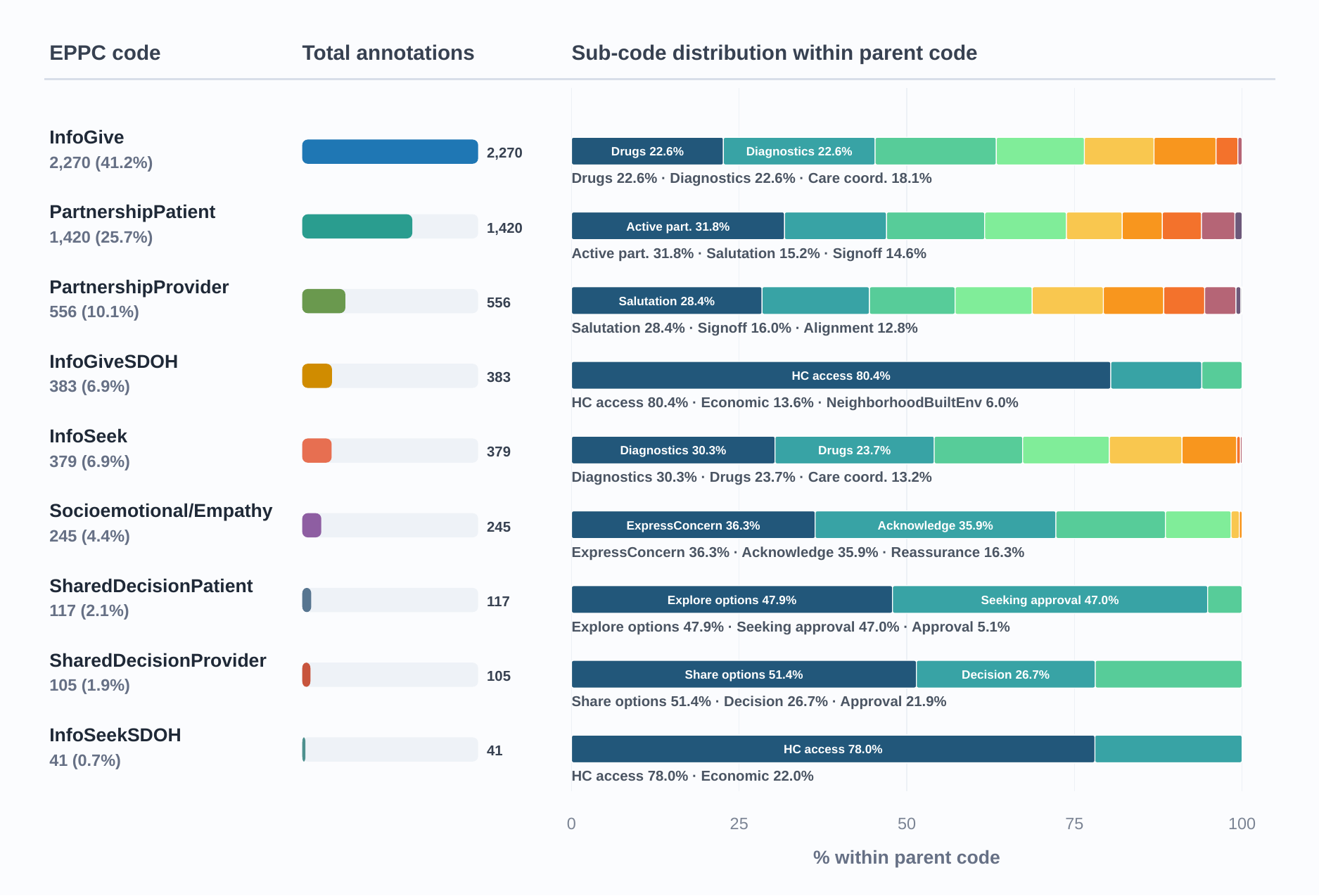}
\caption{
Hierarchical distribution of EPPC code and sub-code annotations. The left panel shows parent-code prevalence across all 5,516 annotations. The right panel shows within-parent sub-code composition.
}
\label{fig:eppc-hierarchy-distribution}
\end{figure*}

The final dataset contained 867 annotated examples and 5,516 EPPC annotations spanning 9 high-level codes, 34 sub-codes, and 52 observed code/sub-code combinations. We used a 70:30 stratified train-test split for model development and evaluation. For complete details regarding the dataset creation methodology including IRB approval, waiver of consent, annotator adjudication protocols, de-identification procedures, and patient-level statistics, we direct readers to the foundational benchmark study \cite{fodeh2026eppcminerben}. Table~\ref{tab:dataset-characteristics} summarizes the dataset characteristics, and Figure~\ref{fig:eppc-hierarchy-distribution} shows the hierarchical distribution of high-level EPPC codes and their observed sub-codes.

% \placeholdertable{Dataset characteristics for the current EPPC extraction dataset.}{tab:dataset-characteristics}{
% Train split & TODO: examples, messages, triples & Dataset summary script & TODO \\
% Test split & TODO: examples, messages, triples & Dataset summary script & TODO \\
% Label inventory & TODO: codes, sub-codes, observed pairs & EPPC ontology files & TODO \\
% Evidence spans & TODO: length distribution & Dataset summary script & TODO \\
% }

\begin{table}[t]
\caption{Dataset characteristics for the EPPC extraction corpus.}
\label{tab:dataset-characteristics}
\centering
\footnotesize
\begin{tabular}{@{}>{\raggedright\arraybackslash}p{0.54\columnwidth}>{\raggedright\arraybackslash}p{0.36\columnwidth}@{}}
\toprule
\textbf{Characteristic} & \textbf{Value} \\
\midrule
\rowcolor{gray!10}\multicolumn{2}{@{}l}{\textbf{Corpus size}} \\
Annotated examples & 867 \\
EPPC annotations & 5,516 \\
Annotations per example & 6.36 mean; 5 median \\
\addlinespace[2pt]
\rowcolor{gray!10}\multicolumn{2}{@{}l}{\textbf{Label inventory}} \\
High-level codes & 9 \\
Sub-codes & 34 \\
Observed code/sub-code pairs & 52 \\
\addlinespace[2pt]
\rowcolor{gray!10}\multicolumn{2}{@{}l}{\textbf{Label distribution}} \\
Most frequent high-level code & InfoGive: 2,270 (41.2\%) \\
Least frequent high-level code & InfoSeekSDOH: 41 (0.7\%) \\
Most frequent sub-code & Diagnostics: 629 (11.4\%) \\
Least frequent sub-code & Sadness/Fear: 3 (0.1\%) \\
\addlinespace[2pt]
\rowcolor{gray!10}\multicolumn{2}{@{}l}{\textbf{Evidence and evaluation split}} \\
Evidence span length & 3.96 mean; 4 median; IQR 2--5 words \\
Train-test split & 70:30 stratified; 607 train / 260 test examples \\
\bottomrule
\end{tabular}
\end{table}

\subsection{Baseline and Comparison Methods}
\label{subsec:baselines}

To contextualize the performance of EPPC-OASIS, we compared methods along two axes: how the model was adapted to the EPPC task and how predictions were produced at inference time. Prompting baselines, including zero-shot and few-shot variants, assessed how well instruction-tuned language models could apply the EPPC schema from task instructions and examples alone. Supervised adaptation baselines measured the benefit of learning directly from annotated EPPC examples, including standard supervised fine-tuning and token-level preference optimization methods using the same input-output format \cite{rafailov2024dpo,meng2024simpo}. We also included STaR-DRO as a prior robustness-oriented comparison method under the same extraction setting \cite{fodeh2026stardro}. The final comparison then evaluated structured inference-refinement procedures applied to adapted models, allowing us to separate improvements from task-specific training, ontology-aware representation learning, and inference-time correction or aggregation.

We evaluated trainable methods on instruction-tuned open-weight models selected to reflect realistic deployment and scaling regimes for clinical NLP research. The primary Llama and Mistral model set included Llama-3.2-1B-Instruct, Llama-3.2-3B-Instruct, Llama-3.1-8B-Instruct, Mistral-Small-24B-Instruct, and Llama-3.3-70B-Instruct \cite{grattafiori2024llama3,meta2024llama32card,meta2024llama33card,mistral2025smallcard}. These models span lightweight local experimentation, mid-scale task-specific adaptation, and high-capacity reference settings. To assess whether the observed patterns were specific to a single model family, the final model-scale analysis also included Qwen2.5-Instruct models at 3B, 7B, and 32B parameters and DeepSeek-R1-Distill-Llama-70B \cite{yang2025qwen25,deepseek2025r1}. This expanded set allowed us to evaluate and validate whether ontology-aware adaptation and structured inference refinement generalized across model families with different instruction-tuning and reasoning behavior.

\subsection{EPPC-OASIS}
\label{subsec:eppc-oasis}

Standard supervised fine-tuning treats EPPC extraction primarily as example-level sequence prediction: the model is optimized to reproduce each target annotation, but the loss does not directly represent relationships among EPPC codes, sub-codes, or valid parent-child label combinations. This is limiting for EPPC coding because the schema is hierarchical, multi-label, and semantically uneven: some labels are closely related, some share sub-code names across different parent codes, and some are rare but clinically meaningful. As a result, prediction errors are often structured rather than arbitrary, such as assigning a plausible sub-code under the wrong parent code or confusing communication behaviors that are semantically adjacent in the ontology. EPPC-OASIS was designed to make this label structure available during training, rather than using the ontology only as a fixed output vocabulary or post hoc evaluation reference.

EPPC-OASIS augments supervised adaptation with an ontology-aware optimal-transport training signal tailored to the EPPC schema \cite{cuturi2013sinkhorn}. Rather than treating each target annotation as an isolated sequence, the method represents each training example in terms of the EPPC codes and sub-codes it contains. These ontology representations define a target neighborhood over memory-bank examples, and the model is trained so that its representation-neighborhood distribution is aligned with that ontology-derived distribution. The following sections describe how ontology representations are constructed, how label-level similarity is defined, and how this signal is incorporated into training with Wasserstein memory-bank alignment.

\subsubsection{Ontology Vector Construction}
\label{subsubsec:ontology-vectors}

EPPC-OASIS first converts the gold annotation for each training example into an ontology-derived vector that can be used as a label-level representation during training. Let \(\mathcal{C}\) denote the set of high-level EPPC codes and \(\mathcal{S}\) the set of EPPC sub-codes. For an example \(x_i\), we define a multi-hot vector

\begin{equation}
\mathbf{o}_i \in \{0,1\}^{|\mathcal{C}| + |\mathcal{S}|}
\label{eq:ontology-vector}
\end{equation}

%\[
%\mathbf{o}_i \in \{0,1\}^{|\mathcal{C}| + |\mathcal{S}|},
%\]
where entries are set to one for every code \(c \in \mathcal{C}\) and sub-code \(s \in \mathcal{S}\) appearing in the gold annotation for that example. When an example contains multiple EPPC annotations, the vector represents the union of all codes and sub-codes expressed in the instance. This vector serves as a multi-hot encoding of active communication behaviors, providing the interface used to compare examples in the subsequent similarity and alignment steps.

\subsubsection{Ontology Similarity Prior}
\label{subsubsec:ontology-prior}

While the multi-hot vectors identify which EPPC labels are present in each example, they do not by themselves capture the hierarchical or graded relationships among labels in the schema. We therefore reintroduce the structure of the EPPC ontology by defining an ontology target similarity between examples using their multi-hot vectors and a label-similarity prior. Let \(\mathbf{o}_i\) and \(\mathbf{o}_j\) be the multi-hot vectors for examples \(x_i\) and \(x_j\), and let \(\mathbf{P}\) denote a nonnegative prior matrix over the combined code and sub-code inventory. The target similarity is defined as

\begin{equation}
t_{ij}
=
\mathrm{clip}_{[0,1]}
\left(
\left( \frac{\mathbf{o}_i}{\|\mathbf{o}_i\|_2} \right)^{\top}
\mathbf{P}
\frac{\mathbf{o}_j}{\|\mathbf{o}_j\|_2}
\right)
\label{eq:ontology-similarity}
\end{equation}

%\[
%t_{ij}
%=
%\mathrm{clip}_{[0,1]}
%\left(
%\frac{\mathbf{o}_i}{\|\mathbf{o}_i\|_2}^{\top}
%\mathbf{P}
%\frac{\mathbf{o}_j}{\|\mathbf{o}_j\|_2}
%\right).
%\]
By defining the similarity through the prior matrix \(\mathbf{P}\), which encodes known relations between codes and sub-codes, we effectively transform the flat label overlap into a structured similarity measure that respects the EPPC schema's organization. When \(\mathbf{P}\) is the identity matrix, this reduces to normalized overlap between active EPPC labels. In the main ontology-aware runs, \(\mathbf{P}\) was constructed from semantic similarity among short textual descriptions of the EPPC codes and sub-codes and used as a smoothing prior over the gold ontology vectors. Thus, the prior can increase the target similarity between related labels, but it does not replace the supervised extraction objective or determine the gold labels for any example. Setting \(\mathbf{P}\) to the identity matrix removes this smoothing while preserving the rest of the ontology-alignment framework.

\subsubsection{Wasserstein Memory-Bank Alignment}
\label{subsubsec:memory-bank}

A practical challenge in applying ontology alignment to EPPC extraction is that useful comparison examples are sparse within ordinary minibatches. The EPPC label distribution is imbalanced, examples may contain multiple code/sub-code annotations, and related label combinations may occur infrequently. This problem becomes more pronounced when fine-tuning larger models, where memory constraints reduce the per-device batch size and make batch-local alignment estimates unstable. Because the EPPC training set is modest in size, the final EPPC-OASIS configuration uses a prefilled frozen memory bank sized to hold the training examples, providing each example with a broader and more stable set of ontology-labeled representations for alignment. 

This design choice addresses the potential "moving target" instability of updating the memory bank during fine-tuning: as the model adapts to the specific extraction task, its hidden representations can shift rapidly, causing alignment neighborhoods to fluctuate. By anchoring to the frozen representations of the base instruction-tuned model, which preserves a more general semantic organization, the Wasserstein objective acts as a stable structured regularizer during adaptation.

For each training example \(x_i\), we extract a hidden representation \(\mathbf{h}_i\) from the final transformer layer at the end of the input prompt, before generation of the target annotation begins. The memory bank stores pairs \((\mathbf{h}_j, \mathbf{o}_j)\), where \(\mathbf{h}_j\) is the stored representation of another training example and \(\mathbf{o}_j\) is its ontology vector. For each current example and memory-bank example, representation similarity is computed as

\begin{equation}
r_{ij}
=
\left( \frac{\mathbf{h}_i}{\|\mathbf{h}_i\|_2} \right)^{\top}
\frac{\mathbf{h}_j}{\|\mathbf{h}_j\|_2}.
\label{eq:representation-similarity}
\end{equation}

%\[
%r_{ij}
%=
%\left( \frac{\mathbf{h}_i}{\|\mathbf{h}_i\|_2} \right)^{\top}
%\frac{\mathbf{h}_j}{\|\mathbf{h}_j\|_2}.
%\]
EPPC-OASIS then forms two distributions over the memory bank. The representation-neighborhood distribution is
\begin{equation}
p^{\mathrm{rep}}_{ij}
=
\frac{\exp(r_{ij}/\tau)}
{\sum_{\ell=1}^{M}\exp(r_{i\ell}/\tau)}
\label{eq:rep-neighborhood}
\end{equation}

%\[
%p^{\mathrm{rep}}_{ij}
%=
%\frac{\exp(r_{ij}/\tau)}
%{\sum_{\ell=1}^{M}\exp(r_{i\ell}/\tau)},
%\]

where \(M\) is the number of examples in the memory bank and \(\tau\) is the softmax temperature. The ontology-neighborhood distribution is obtained by normalizing the target similarities,
\begin{equation}
p^{\mathrm{ont}}_{ij}
=
\frac{t_{ij}}
{\sum_{\ell=1}^{M}t_{i\ell}+\epsilon_0}
\label{eq:ont-neighborhood}
\end{equation}

%\[
%p^{\mathrm{ont}}_{ij}
%=
%\frac{t_{ij}}
%{\sum_{\ell=1}^{M}t_{i\ell}+\epsilon_0},
%\]

where \(\epsilon_0\) is a small numerical constant. To compare these two neighborhoods, we define a transport cost between memory-bank examples
\begin{equation}
C_{jk}=1-t_{jk}
\label{eq:transport-cost}
\end{equation}

%\[
%C_{jk}=1-t_{jk}.
%\]

The ontology-alignment loss for \(x_i\) is the entropically regularized optimal-transport cost between the representation and ontology distributions,
\begin{equation}
\mathcal{L}_{\mathrm{OT},i}
=
\mathrm{Sinkhorn}_{\epsilon}
\left(
p^{\mathrm{rep}}_{i},
p^{\mathrm{ont}}_{i},
\mathbf{C}
\right)
\label{eq:ot-loss}
\end{equation}

%\[
%\mathcal{L}_{\mathrm{OT},i}
%=
%\mathrm{Sinkhorn}_{\epsilon}
%\left(
%p^{\mathrm{rep}}_{i},
%p^{\mathrm{ont}}_{i},
%\mathbf{C}
%\right),
%\]

where \(\epsilon\) is the Sinkhorn regularization parameter. This Wasserstein formulation aligns the full neighborhood induced by model representations with the neighborhood implied by EPPC ontology similarity, rather than optimizing each similarity relation independently \cite{cuturi2013sinkhorn}.

\subsubsection{Training Objective and Algorithm}
\label{subsubsec:training-objective}

The final EPPC-OASIS objective combines the supervised extraction loss with the Wasserstein ontology-alignment loss. Let \(\mathcal{L}_{\mathrm{SFT}}\) denote the standard autoregressive training loss for generating the gold structured annotation. The ontology loss is the mean optimal-transport contribution across training examples,
\begin{equation}
\mathcal{L}_{\mathrm{ont}}
=
\frac{1}{N}
\sum_i \mathcal{L}_{\mathrm{OT},i}
\label{eq:ontology-loss}
\end{equation}

The total training objective is

\begin{equation}
\mathcal{L}
=
\mathcal{L}_{\mathrm{SFT}}
+
\lambda_{\mathrm{ont}}
\mathcal{L}_{\mathrm{ont}}
\label{eq:total-loss}
\end{equation}

%\[
%\mathcal{L}_{\mathrm{ont}}
%=
%\frac{1}{N}
%\sum_i \mathcal{L}_{\mathrm{OT},i}.
%\]
%The total training objective is
%\[
%\mathcal{L}
%=
%\mathcal{L}_{\mathrm{SFT}}
%+
%\lambda_{\mathrm{ont}}
%\mathcal{L}_{\mathrm{ont}},
%\]

where \(\lambda_{\mathrm{ont}}\) controls the strength of ontology alignment. The ontology term is applied only after the memory bank contains enough labeled representations to provide stable neighborhoods. Thus, EPPC-OASIS preserves the supervised extraction objective as the primary learning signal while using Wasserstein ontology alignment as a structured regularizer during fine-tuning.

\begin{algorithm}[t]
\caption{EPPC-OASIS training loop}
\label{alg:eppc-oasis}
\begin{algorithmic}[1]
\Require Training examples, EPPC label inventory, optional prior matrix \(\mathbf{P}\), base language model
\State Construct ontology vector \(\mathbf{o}_i\) for each training example from gold EPPC codes and sub-codes
\State Prefill the frozen memory bank with training-example representations and ontology vectors
\For{each training minibatch}
    \State Compute supervised extraction loss \(\mathcal{L}_{\mathrm{SFT}}\)
    \State Extract prompt-level hidden representations \(\mathbf{h}_i\)
    \State Compute representation similarities \(r_{ij}\) and ontology target similarities \(t_{ij}\)
    \State Construct \(p_i^{\mathrm{rep}}\), \(p_i^{\mathrm{ont}}\), and transport costs \(C_{jk}=1-t_{jk}\)
    \State Compute Sinkhorn ontology-alignment loss \(\mathcal{L}_{\mathrm{ont}}\)
    \State Update model using \(\mathcal{L}_{\mathrm{SFT}}+\lambda_{\mathrm{ont}}\mathcal{L}_{\mathrm{ont}}\)
\EndFor
\end{algorithmic}
\end{algorithm}

\subsection{Structured Inference Refinement}
\label{subsec:structured-inference}

EPPC-OASIS changes the adapted model, but structured extraction errors can still arise at inference time because label assignment and evidence grounding fail in different ways. A greedy output may preserve an appropriate evidence span while assigning an adjacent sub-code, whereas a sampled or refinement-augmented output may recover a better label set while drifting from the source text. We therefore evaluated a structured inference-refinement layer on top of the adapted models. This layer does not change the training objective, instead, it applies reproducible inference procedures that verify, resample, merge, or select structured predictions before final scoring.

\subsubsection{Chain-of-Thought Self-Refinement(CoT-SR)}
\label{subsubsec:self-refinement}

The CoT-SR inference pipeline begins with greedy structured generation and then performs targeted verification of the predicted annotations, following chain-of-thought self-verification and self-refinement strategies for test-time correction \cite{wei2023cot,madaan2023selfrefine}. Stage 1 generates a greedy structured JSON output. Stage 2 applies a local span check: a span is accepted if it appears verbatim in the source sentence or if its token-level Jaccard overlap with the sentence is at least 0.8; examples with any remaining invalid spans are passed through a single greedy verification prompt that asks the model to remove unsupported triplets, fix minor span errors, and retain Code/Sub-code assignments consistent with the EPPC schema. Stage 3 runs a separate self-refinement prompt on all verified outputs, generating multiple refinement samples; the refined candidate is selected by majority voting over structured triplets with tie-breaking that favors context-supported spans. Stage 4 selects per example, the better of the verified and refined outputs using span-validity criteria. Only the final structured JSON is scored; intermediate generation text is not used as the prediction. This procedure is intended to improve grounded structured extraction without treating intermediate reasoning text as part of the final annotation.

%The CoT-SR inference pipeline begins with greedy structured generation and then performs targeted verification of the predicted annotations, following the broader use of chain-of-thought self-verification and self-refinement strategies for test-time correction \cite{wei2023cot,madaan2023selfrefine}. The verification step checks whether predicted spans are supported by the original message context and whether the Code and Sub-code assignments remain consistent with the EPPC schema. Predictions requiring correction are passed through a refinement prompt that asks the model to keep correct triplets, remove unsupported triplets, fix minor span errors when the intended evidence is clear, and add missed annotations only when supported by the source text. Multiple refinement samples can be generated, after which the selected output favors predictions with valid evidence spans and a coherent number of structured annotations. This procedure is intended to improve grounded structured extraction without treating intermediate reasoning text as part of the final prediction.

\subsubsection{Self-Consistency and Label-Corrected Hybrid Prediction}
\label{subsubsec:self-consistency-hybrid}

Self-consistency inference generates multiple sampled structured outputs for the same input and aggregates them by voting over predicted Code/Sub-code pairs \cite{wang2023selfconsistency}. A (Code, Sub-code) pair is retained if it appears in at least half of the N samples; for each retained pair, the final span is chosen as the candidate with the highest mean token-Jaccard similarity to all other sampled spans for that pair. This procedure primarily targets label stability, since repeated sampling can reveal which EPPC categories are consistently recovered even when individual spans vary. We also evaluated label-corrected hybrid prediction, which combines the complementary strengths of greedy and self-consistency outputs. Each greedy triplet retains its evidence span; for each greedy span, the self-consistency span with the highest token Jaccard is identified, and if that overlap is at least 0.5, the greedy Code and Sub-code labels are replaced with the self-consistency labels while the greedy span is kept. Self-consistency triplets whose (Code, Sub-code) pair is not already represented are then appended with their original self-consistency spans. The mild variant is used throughout, which keeps greedy labels unchanged when no self-consistency span exceeds the Jaccard threshold.

%Self-consistency inference generates multiple sampled structured outputs for the same input and aggregates them by voting over predicted Code/Sub-code pairs \cite{wang2023selfconsistency}. For each pair supported by a sufficient fraction of samples, the final span is selected from the sampled spans with the strongest within-sample agreement. This procedure primarily targets label stability, since repeated sampling can reveal which EPPC categories are consistently recovered even when individual spans vary. We also evaluated label-corrected hybrid prediction, which combines the complementary strengths of greedy and self-consistency outputs. In this hybrid, the greedy or refined output supplies evidence spans, while self-consistency supplies more stable Code/Sub-code labels when the sampled predictions agree on labels associated with similar spans. Additional self-consistency triplets may be added when they represent label pairs not covered by the greedy output.

\subsubsection{Selector and Ensemble Variants}
\label{subsubsec:selector-ensemble}

For some models, no single inference stage dominated all annotation components. We therefore evaluated deployable selector and ensemble variants that choose or merge complete predictions using only prediction-internal criteria, rather than gold labels. The selector ensemble compares CoT-SR and self-consistency outputs for each example and selects the prediction with more verbatim context-supported spans; when tied, it favors the output with more distinct (Code, Sub-code) pairs. For high-capacity models with multiple trained seeds, we also evaluated soft-majority merging of greedy outputs across multiple trained seeds by grouping triplets that share the same (Code, Sub-code) pair and have span token Jaccard similarity of at least 0.5, retaining groups supported by at least two distinct seeds and using the longest span in each group as the representative. These approaches are reported only when they define a single deployable prediction pipeline. Component-wise best Code, Sub-code, or Span scores from different inference stages are treated as diagnostic upper bounds rather than primary method results.

\subsubsection{Coverage-Guided Resampling(CGRA) \& Span Reranking}
\label{subsubsec:coverage-span-variants}

We also considered inference procedures designed for specific structured-output failure modes. Coverage-guided resampling (CGRA) begins from greedy decoding, flags examples with fewer than four predicted triplets, generates additional sampled outputs only for those examples, and augments the baseline prediction with any new triplet whose span appears verbatim in the source context. This targets under-extraction without applying multi-sample inference uniformly to the full test set. The span-anchored ontology reranker combines candidate predictions from multiple inference stages, including greedy decoding, CoT-SR, and self-consistency, by fuzzy-matching predicted labels against the EPPC ontology, snapping near-miss spans to the source context when a single high-confidence alignment exists (token Jaccard at least 0.72 with a margin of 0.08 over the next-best candidate), and scoring triplets by weighted source agreement with a penalty for snapped spans. These variants were included in the inference-method inventory and are reported in the main tables only when the final prediction file corresponds to a single deployable pipeline; otherwise, they are used as diagnostic analyses of label coverage, span support, and source disagreement.

% \placeholderfigure{Two-stage EPPC extraction framework. The final figure should show ontology-aware fine-tuning followed by structured inference refinement, including greedy generation, self-verification/refinement, self-consistency voting, label-corrected hybrid prediction, and selector or seed-merge variants.}{fig:two-stage-framework}{Placeholder for Figure 4: two-stage EPPC-OASIS and structured inference framework}

\subsection{Evaluation Metrics}
\label{subsec:evaluation}

Evaluation was organized to preserve comparability with prior EPPC extraction work while also reporting stricter metrics for the structured prediction target used in this study. Following earlier evaluations, we report Code F1, Sub-code F1, and Span F1, where span matching treats a predicted evidence span as correct when its token-level Jaccard overlap with a gold span is at least 0.6. This threshold was selected because exact token matching is overly brittle for clinical text, where models may validly include or exclude surrounding punctuation or short function words without altering clinical meaning. These metrics describe performance on the individual components of an annotation, but they do not require the components to be correct together. We therefore also report Code+Sub-code F1 as the main label-based metric, requiring the parent code and sub-code to match as a pair, and Triplet F1 as the main grounded extraction metric, requiring the Code, Sub-code, and evidence span match jointly. 

All metrics are micro-averaged across the test set to reflect global corpus-level performance given the severe class imbalance, meaning performance on rare labels contributes proportionally to their frequency. During evaluation, duplicate predicted triplets (identical Code, Sub-code, and Span combinations within the same example) are deduplicated prior to scoring to prevent inflated recall through redundant generation. To avoid one-directional overlap exploitation, all matching was computed symmetrically between predicted and gold annotations using a greedy 1:1 alignment strategy ranked by label correctness and maximum span overlap.

For statistical reporting, final test-set results were summarized across repeated runs with different random seeds. For each model and method, we report the mean and standard deviation of each evaluation metric across seeds, allowing performance differences to be interpreted relative to run-to-run variation during fine-tuning and decoding. Malformed outputs that could not be recovered into the target schema were counted as invalid predictions and contributed no matched annotations. Predicted codes or sub-codes outside the EPPC ontology were treated as unmatched labels for the corresponding label-level and triplet-level comparisons.

%In addition to the primary extraction metrics, we report diagnostic measures including invalid-output rate, invalid-label rate, and label-level error patterns to characterize whether performance differences reflect improved EPPC classification, improved schema adherence, or changes in the distribution of extraction errors.

%
% TODO: Compute and report diagnostic metrics for final runs.
% - Invalid JSON rate: percentage of outputs that cannot be parsed into the target schema.
% - Invalid label rate: percentage of predicted codes/sub-codes outside the EPPC ontology.
% - Difficulty-score analysis: whether examples assigned higher EPPC-OASIS difficulty weights actually have lower accuracy.
% - Per-code or per-sub-code F1: whether performance gains are concentrated in rare labels or common labels.
% - Span-boundary error rate: how often the labels are correct but evidence grounding fails.
% - Over-extraction rate: how often the model predicts too many annotations relative to gold.

\subsection{Implementation Details}
\label{subsec:implementation-details}

Experiments were conducted as single-GPU runs on institutional accelerator nodes equipped with NVIDIA B200 or H200 GPUs, with the specific device determined by model size and scheduler availability. Trainable open-weight models were adapted using parameter-efficient fine-tuning with LoRA adapters \cite{hu2021lora}. For larger models, adapters were trained with 4-bit quantized model loading, following the practical QLoRA fine-tuning regime, to make repeated seed runs feasible without changing the underlying training objective \cite{dettmers2023qlora}. 

The implementation used PyTorch and the Hugging Face ecosystem, including Transformers, PEFT, bitsandbytes, Unsloth for efficient adapter-based fine-tuning, and POT for Sinkhorn optimal transport; vLLM was used for batched inference and for multi-sample inference procedures \cite{paszke2019pytorch,wolf2020transformers,mangrulkar2022peft,dettmers2021optimizers,unsloth2024software,flamary2021pot,kwon2023vllm}. Ontology-vector construction, ontology-prior generation, Wasserstein memory-bank alignment, output parsing, and metric computation were applied consistently across all evaluated methods. Final runs used fixed random seeds, common parser-recovery settings, and the same span-matching threshold defined above.

%Software versions, command-line configurations, prediction files, and non-PHI experiment artifacts will be released to support reproducibility of the reported analyses.

Training hyperparameters were selected by model scale and then held fixed across repeated-seed runs for each model-method configuration. All trainable methods used LoRA adapters with rank 32, LoRA scaling factor 64, LoRA dropout of 0.05, the 8-bit AdamW optimizer, a cosine learning-rate schedule, and an effective batch size of 16. Repeated runs varied only the random seed. For EPPC-OASIS, the memory bank was sized to hold the training set, the ontology-alignment term was applied without a separate warmup period, and final Wasserstein runs used a prefilled frozen memory bank with Sinkhorn regularization of 0.01. Model-specific learning rates, ontology-loss weights, and Sinkhorn regularization values are provided in Appendix Table~\ref{tab:training-hyperparameters}.

We also evaluated a weight-averaged EPPC-OASIS checkpoint constructed from independently trained random-seed runs. Weight averaging is motivated by the observation that independently fine-tuned checkpoints with similar validation behavior can occupy a connected low-loss region of parameter space; averaging can move the resulting model toward the center of this region, reducing seed-specific optimization noise while preserving single-checkpoint inference. We therefore averaged the independently trained EPPC-OASIS checkpoints parameter-wise and evaluated the resulting checkpoint with the same greedy decoding protocol as the individual seed runs.

Structured inference settings were fixed within each reported pipeline. Greedy inference used temperature 0.0. Self-consistency used multiple sampled outputs per example, with sample counts and temperatures selected before final reporting for each model family and documented with the corresponding prediction files. CoT-SR used greedy generation followed by verification and refinement passes, with refinement sample counts documented in the run configuration. Selector, label-corrected hybrid, coverage-guided resampling, span-anchored reranking, and seed-merged variants were reported only when they produced a single complete prediction file for each test example. For each reported row, we recorded the training checkpoint source separately from the inference pipeline so that post-training improvements were not attributed to EPPC-OASIS unless the underlying checkpoint used the EPPC-OASIS training objective. Component-wise best scores obtained by taking Code, Sub-code, and Span values from different inference stages were not treated as primary deployable results.

\section{Results}
\label{sec:results}

% \subsection{Dataset Characteristics}
% \label{subsec:results-dataset}

% Table~\ref{tab:dataset-characteristics} summarizes the final EPPC extraction dataset. The dataset is relatively small in number of examples but dense in annotation, with multiple EPPC labels often assigned within the same message context. The label distribution is also markedly imbalanced: a small number of high-frequency communication categories account for a large fraction of annotations, while several clinically relevant behaviors occur sparsely. This combination of dense structured annotation and long-tailed label frequency makes the task more demanding than isolated single-label classification, because models must learn frequent EPPC behaviors without ignoring rare code/sub-code combinations.

\subsection{Training and Inference Strategy Comparison}
\label{subsec:main-results}

The primary method comparison used Llama-3.1-8B-Instruct as the reference open-weight model, allowing differences among training and inference strategies to be interpreted without confounding from model scale. Table~\ref{tab:main-results} compares prompting-only baselines, supervised fine-tuning, EPPC-OASIS with greedy inference, and deployable structured inference-refinement variants under the same train-test split, output schema, parser, and evaluation metrics. The preference-optimization baselines included Direct Preference Optimization (DPO), Identity Preference Optimization (IPO), Simple Preference Optimization (SimPO), and Odds Ratio Preference Optimization (ORPO) as well as Stateful Tsallis Reweighting for Distributionally Robust Optimization (STaR-DRO) as a robustness-oriented baseline. Prompting baselines measure how much of the EPPC schema can be applied from instructions and in-context examples alone. The trainable baselines measure the effect of task-specific adaptation, while the inference-refinement rows test whether self-verification, self-consistency, and hybrid prediction further improve structured extraction after adaptation.

\begin{table*}[t]
\caption{Training and inference strategy comparison using Llama-3.1-8B-Instruct as the reference model. Rows correspond to deployable pipelines that produce one complete prediction for each test example. Values are reported as mean (standard deviation) across random seeds for trainable methods when repeated runs are available.}
\label{tab:main-results}
\centering
\scriptsize
\setlength{\tabcolsep}{3pt}
\begin{tabular}{@{}>{\raggedright\arraybackslash}p{0.17\textwidth}>{\raggedright\arraybackslash}p{0.18\textwidth}ccccc@{}}
\toprule
\makecell{\textbf{Training}\\\textbf{checkpoint}} & \makecell{\textbf{Inference}\\\textbf{pipeline}} & \makecell{\textbf{Code}\\\textbf{F1 (\%) \(\uparrow\)}} & \makecell{\textbf{Sub-code}\\\textbf{F1 (\%) \(\uparrow\)}} & \makecell{\textbf{Span}\\\textbf{F1 (\%) \(\uparrow\)}} & \makecell{\textbf{Code+Sub-code}\\\textbf{F1 (\%) \(\uparrow\)}} & \makecell{\textbf{Triplet}\\\textbf{F1 (\%) \(\uparrow\)}} \\
\midrule
Prompting only & Zero-shot greedy & 35.36 & 14.92 & 35.78 & 5.95 & 3.03 \\
Prompting only & Few-shot greedy & 58.08 & 44.00 & 75.56 & 26.63 & 16.29 \\
\midrule
Supervised fine-tuning & Greedy & 84.66 \ensuremath{\pm} 0.70 & 77.83 \ensuremath{\pm} 0.79 & \bestresult{83.28 \ensuremath{\pm} 0.72} & 74.93 \ensuremath{\pm} 0.11 & 60.36 \ensuremath{\pm} 1.05 \\
\midrule
DPO & Greedy & 84.27 \ensuremath{\pm} 0.23 & 77.83 \ensuremath{\pm} 0.19 & 83.25 \ensuremath{\pm} 0.03 & 74.77 \ensuremath{\pm} 0.18 & \bestresult{60.81 \ensuremath{\pm} 0.25} \\
IPO & Greedy & 84.21 \ensuremath{\pm} 0.05 & 77.83 \ensuremath{\pm} 0.12 & 83.29 \ensuremath{\pm} 0.04 & 74.71 \ensuremath{\pm} 0.13 & 60.58 \ensuremath{\pm} 0.13 \\
SimPO & Greedy & 84.28 \ensuremath{\pm} 0.07 & 77.81 \ensuremath{\pm} 0.05 & 83.25 \ensuremath{\pm} 0.10 & 74.70 \ensuremath{\pm} 0.11 & 60.85 \ensuremath{\pm} 0.15 \\
ORPO & Greedy & 84.28 \ensuremath{\pm} 0.03 & 77.75 \ensuremath{\pm} 0.03 & \bestresult{83.36 \ensuremath{\pm} 0.04} & 74.59 \ensuremath{\pm} 0.07 & 60.72 \ensuremath{\pm} 0.10 \\
\midrule
STaR-DRO & Greedy & 84.87 \ensuremath{\pm} 0.30 & 77.33 \ensuremath{\pm} 0.18 & 82.95 \ensuremath{\pm} 0.33 & 73.82 \ensuremath{\pm} 0.16 & 59.69 \ensuremath{\pm} 0.37 \\
\midrule
EPPC-OASIS & Greedy & 86.07 \ensuremath{\pm} 0.92 & \bestresult{78.00 \ensuremath{\pm} 0.18} & 82.85 \ensuremath{\pm} 0.48 & \bestresult{75.51 \ensuremath{\pm} 1.00} & 60.57 \ensuremath{\pm} 0.59 \\
\midrule
\multirow{5}{=}{EPPC-OASIS} & CoT-SR & \bestresult{86.47 \ensuremath{\pm} 0.74} & 77.84 \ensuremath{\pm} 0.37 & 82.79 \ensuremath{\pm} 0.68 & 74.68 \ensuremath{\pm} 0.74 & 60.65 \ensuremath{\pm} 0.90 \\
& Self-consistency & 86.31 \ensuremath{\pm} 0.22 & 78.11 \ensuremath{\pm} 0.60 & 69.96 \ensuremath{\pm} 0.76 & 74.75 \ensuremath{\pm} 0.36 & 53.61 \ensuremath{\pm} 0.42 \\
& Label-corrected hybrid & \bestresult{86.47 \ensuremath{\pm} 0.74} &\bestresult{78.24 \ensuremath{\pm} 0.51} & 82.87 \ensuremath{\pm} 0.61 & 74.98 \ensuremath{\pm} 0.51 & 60.41 \ensuremath{\pm} 0.81 \\
& Selector ensemble & 86.47 \ensuremath{\pm} 0.68 & 77.89 \ensuremath{\pm} 0.46 & 82.75 \ensuremath{\pm} 0.67 & 74.69 \ensuremath{\pm} 0.76 & 60.66 \ensuremath{\pm} 0.97 \\
& CGRA & 86.09 \ensuremath{\pm} 1.19 & 77.19 \ensuremath{\pm} 0.49 & 80.41 \ensuremath{\pm} 0.78 & 73.93 \ensuremath{\pm} 0.69 & 59.40 \ensuremath{\pm} 1.11 \\
\bottomrule
\end{tabular}
\end{table*}

The comparison separates the contribution of ontology-aware training from the contribution of structured inference. On the Llama-3.1-8B-Instruct reference model, EPPC-OASIS with greedy inference improved Code F1 from 84.66 to 86.07, Code+Sub-code F1 from 74.93 to 75.51, and Triplet F1 from 60.36 to 60.57 relative to supervised fine-tuning, indicating the effect of the ontology-aware adaptation objective before any inference-time refinement. 

The structured inference variants did not uniformly improve the primary deployable metrics: self-consistency improved Sub-code F1 but substantially reduced span and Triplet F1, while selector-style aggregation produced the highest Triplet F1 among the EPPC-OASIS inference rows at 60.66. Prompting-only baselines showed that providing in-context examples (few-shot) provided a substantial boost over zero-shot instructions, but remained substantially lower on the joint metrics than any adapted model. EPPC-OASIS achieved the highest structural label coherence (Code+Sub-code F1) among all adaptation strategies, consistently outperforming both standard supervised fine-tuning and the robustness-oriented STaR-DRO baseline on structural metrics and grounded Triplet F1. 

Preference-optimization baselines, which operate directly on token sequences, were highly competitive for exact span localization (Span F1) and consequently achieved slightly higher Triplet F1 scores. This highlights a clear trade-off: EPPC-OASIS best preserves the complex label hierarchy, while span boundary generation remains a distinct challenge. These results support the central framing of the study: ontology-aware adaptation improves the structural label backbone, while inference refinement is required to close the gap on evidence grounding.

\begin{figure*}[t]
\centering
\includegraphics[width=\textwidth]{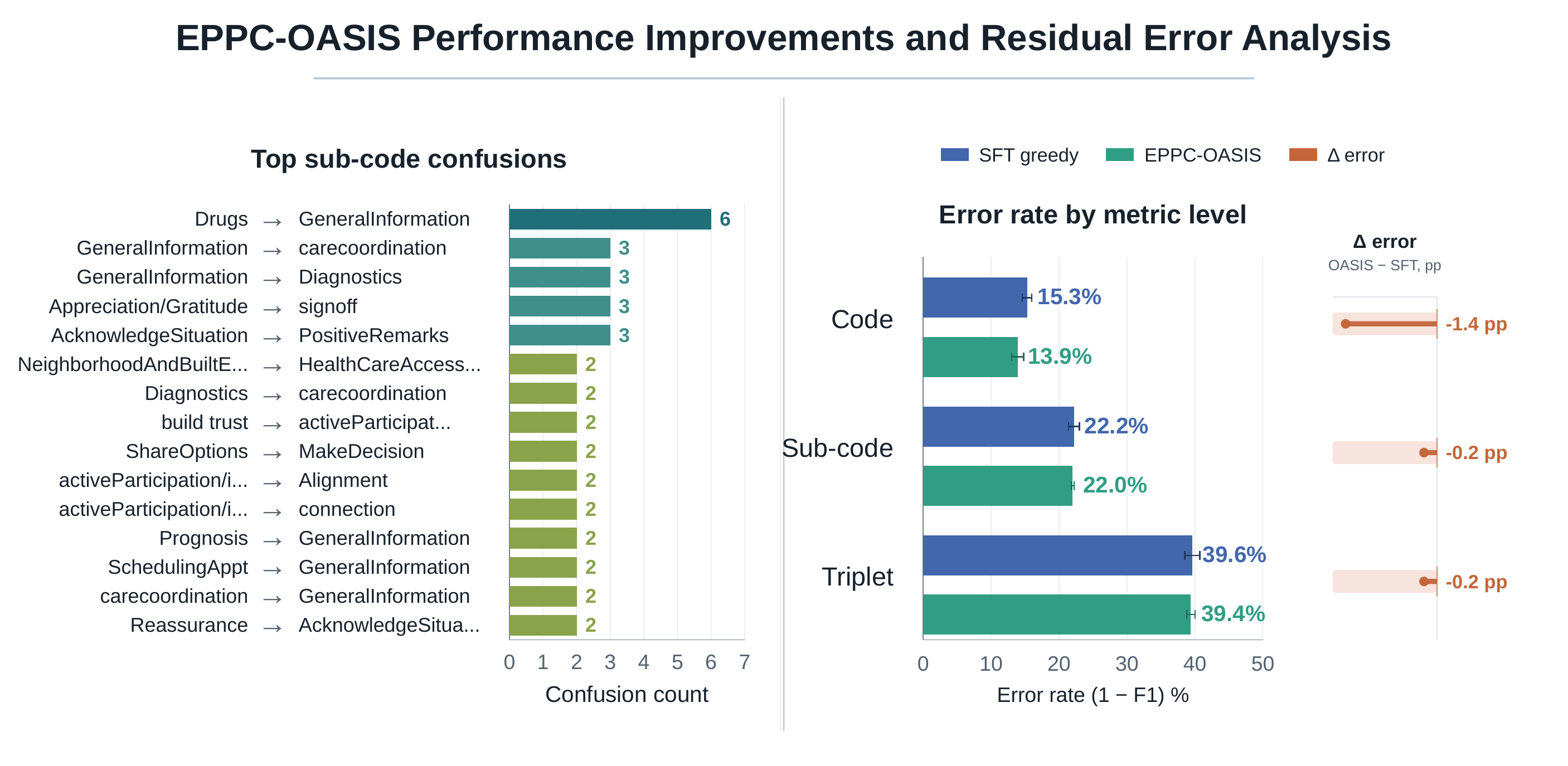}
\caption{
EPPC-OASIS performance improvements and residual sub-code confusions using the Llama-3.1-8B reference model. The left panel shows the most frequent remaining sub-code confusions after EPPC-OASIS greedy inference. Most residual confusions occur between semantically adjacent EPPC sub-codes. The right panel compares extraction error rates between supervised fine-tuning (SFT greedy) and EPPC-OASIS greedy across Code, Sub-code, and Triplet evaluation levels. Lower error rates indicate improved structured extraction performance.
}
\label{fig:eppc_oasis_error_analysis}
\end{figure*}

Figure~\ref{fig:eppc_oasis_error_analysis} further summarizes this pattern for the reference model by pairing the aggregate error-rate reductions with the residual sub-code confusions that remain after ontology-aware adaptation.

%
% \placeholdertable{Main comparison on the current EPPC extraction test set.}{tab:main-results}{
% Zero-shot prompting & TODO: metrics & Final rerun table & TODO \\
% Few-shot prompting & TODO: metrics & Final rerun table & TODO \\
% Supervised fine-tuning & TODO: metrics & Final rerun table & TODO \\
% STaR-DRO & TODO: metrics & Final rerun table & TODO \\
% EPPC-OASIS & TODO: metrics & Final rerun table & TODO \\
% }

\subsection{Performance Across Model Families and Scale}
\label{subsec:scaling-results}

We evaluated supervised fine-tuning and EPPC-OASIS across multiple instruction-tuned model scales to determine whether ontology-aware adaptation generalized beyond a single backbone. Table~\ref{tab:scaling-results} reports results for the available model families using the same test set, output schema, parser, and evaluation metrics. For each model, the table records the SFT greedy baseline, the corresponding EPPC-OASIS greedy result, and a weight-averaged EPPC-OASIS checkpoint when available. The weight-averaged rows evaluate whether the variation across EPPC-OASIS seed runs can be consolidated into a single deployable checkpoint. EPPC-OASIS improved Triplet F1 for all evaluated Llama and Mistral scales. Notably, while greedy EPPC-OASIS results were mixed for the Qwen2.5 family, weight averaging consolidated the training signal to produce consistent gains across all Qwen scales, including a +2.12 Triplet F1 improvement for Qwen2.5-32B-Instruct. Weight averaging further improved the strongest Triplet F1 for the 1B, 3B, 8B, and 24B models. These results suggest that ontology-aware adaptation is most beneficial when the base model has less task-specific capacity, and that weight averaging serves as a robust mechanism for stabilizing these gains across different model architectures.

\begin{table*}[t]
\caption{Performance across model families and scales. All results use greedy decoding. Repeated-seed rows report mean (standard deviation); weight-averaged rows report the single averaged checkpoint.}
\label{tab:scaling-results}
\centering
\scriptsize
\setlength{\tabcolsep}{2.5pt}
\begin{tabular}{@{}>{\raggedright\arraybackslash}p{0.28\textwidth}>{\raggedright\arraybackslash}p{0.18\textwidth}ccccc@{}}
\toprule
\makecell{\textbf{Model}} & \makecell{\textbf{Method}} & \makecell{\textbf{Code}\\\textbf{F1 (\%) \(\uparrow\)}} & \makecell{\textbf{Sub-code}\\\textbf{F1 (\%) \(\uparrow\)}} & \makecell{\textbf{Span}\\\textbf{F1 (\%) \(\uparrow\)}} & \makecell{\textbf{Code+Sub-code}\\\textbf{F1 (\%) \(\uparrow\)}} & \makecell{\textbf{Triplet}\\\textbf{F1 (\%) \(\uparrow\)}} \\
\midrule
Llama-3.2-1B-Instruct & SFT & 83.25 \ensuremath{\pm} 0.59 & 74.42 \ensuremath{\pm} 0.64 & 81.04 \ensuremath{\pm} 0.88 & 70.97 \ensuremath{\pm} 0.99 & 53.99 \ensuremath{\pm} 0.99 \\
& EPPC-OASIS & 83.87 \ensuremath{\pm} 0.96 & 75.22 \ensuremath{\pm} 1.14 & 80.29 \ensuremath{\pm} 0.43 & 72.15 \ensuremath{\pm} 0.74 & 56.35 \ensuremath{\pm} 0.90 \\
& EPPC-OASIS (weight avg.) & \bestresult{84.80} & \bestresult{77.07} & \bestresult{83.54} & \bestresult{73.71} & \bestresult{59.88} \\
\midrule
Llama-3.2-3B-Instruct & SFT & 84.98 \ensuremath{\pm} 0.74 & 77.26 \ensuremath{\pm} 0.30 & 82.45 \ensuremath{\pm} 0.48 & 74.05 \ensuremath{\pm} 0.09 & 58.95 \ensuremath{\pm} 0.31 \\
& EPPC-OASIS & 84.02 \ensuremath{\pm} 0.17 & 76.14 \ensuremath{\pm} 1.40 & 81.49 \ensuremath{\pm} 0.69 & 74.72 \ensuremath{\pm} 0.33 & 59.24 \ensuremath{\pm} 0.94 \\
& EPPC-OASIS (weight avg.) & \bestresult{85.89} & \bestresult{78.38} & \bestresult{82.68} & \bestresult{75.24} & \bestresult{61.79} \\
\midrule
Llama-3.1-8B-Instruct & SFT & 84.66 \ensuremath{\pm} 0.70 & 77.83 \ensuremath{\pm} 0.79 & 83.28 \ensuremath{\pm} 0.72 & 74.93 \ensuremath{\pm} 0.11 & 60.36 \ensuremath{\pm} 1.05 \\
& EPPC-OASIS & 86.07 \ensuremath{\pm} 0.92 & 78.00 \ensuremath{\pm} 0.18 & 82.85 \ensuremath{\pm} 0.48 & 75.51 \ensuremath{\pm} 1.00 & 60.57 \ensuremath{\pm} 0.59 \\
& EPPC-OASIS (weight avg.) & \bestresult{86.76} & \bestresult{78.65} & \bestresult{84.11} & \bestresult{75.56} & \bestresult{62.99} \\
\midrule
Mistral-Small-24B-Instruct & SFT & 85.92 \ensuremath{\pm} 0.75 & 77.82 \ensuremath{\pm} 1.03 & 83.80 \ensuremath{\pm} 1.30 & 74.84 \ensuremath{\pm} 0.98 & 60.86 \ensuremath{\pm} 1.18 \\
& EPPC-OASIS & 85.82 \ensuremath{\pm} 0.63 & 78.39 \ensuremath{\pm} 1.02 & \bestresult{84.45 \ensuremath{\pm} 0.39} & 75.26 \ensuremath{\pm} 0.46 & 60.89 \ensuremath{\pm} 0.55 \\
& EPPC-OASIS (weight avg.) & \bestresult{86.85} & \bestresult{79.30} & 84.19 & \bestresult{75.89} & \bestresult{62.75} \\
\midrule
Llama-3.3-70B-Instruct & SFT & 85.13 \ensuremath{\pm} 0.38 & 78.33 \ensuremath{\pm} 0.19 & 82.73 \ensuremath{\pm} 0.50 & 74.64 \ensuremath{\pm} 0.67 & 60.07 \ensuremath{\pm} 0.79 \\
& EPPC-OASIS & 86.02 \ensuremath{\pm} 1.04 & 77.87 \ensuremath{\pm} 0.79 & \bestresult{82.85 \ensuremath{\pm} 0.33} & 74.91 \ensuremath{\pm} 0.42 & \bestresult{61.22 \ensuremath{\pm} 0.04} \\
& EPPC-OASIS (weight avg.) & \bestresult{87.57} & \bestresult{78.79} & 81.71 & \bestresult{75.33} & 60.97 \\
\midrule
Qwen2.5-3B-Instruct & SFT & 85.28 \ensuremath{\pm} 0.53 & 77.74 \ensuremath{\pm} 0.21 & 82.13 \ensuremath{\pm} 0.53 & 74.29 \ensuremath{\pm} 0.31 & 58.84 \ensuremath{\pm} 0.46 \\
& EPPC-OASIS & 84.95 \ensuremath{\pm} 0.10 & 77.51 \ensuremath{\pm} 0.08 & 82.01 \ensuremath{\pm} 0.32 & 73.80 \ensuremath{\pm} 0.23 & 58.88 \ensuremath{\pm} 0.39 \\
& EPPC-OASIS (weight avg.) & \bestresult{85.76} & \bestresult{79.01} & \bestresult{82.33} & \bestresult{75.17} & \bestresult{59.35} \\
\midrule
Qwen2.5-7B-Instruct & SFT & \bestresult{85.59 \ensuremath{\pm} 0.42} & 77.76 \ensuremath{\pm} 0.21 & 82.95 \ensuremath{\pm} 0.42 & 74.36 \ensuremath{\pm} 0.26 & 59.51 \ensuremath{\pm} 0.10 \\
& EPPC-OASIS & 85.00 \ensuremath{\pm} 0.92 & 77.68 \ensuremath{\pm} 0.57 & 83.18 \ensuremath{\pm} 0.31 & 74.07 \ensuremath{\pm} 0.67 & 59.55 \ensuremath{\pm} 0.63 \\
& EPPC-OASIS (weight avg.) & 85.37 & \bestresult{78.79} & \bestresult{83.52} & \bestresult{74.57} & \bestresult{60.61} \\
\midrule
Qwen2.5-32B-Instruct & SFT & 86.95 \ensuremath{\pm} 0.63 & 78.71 \ensuremath{\pm} 0.47 & 84.09 \ensuremath{\pm} 0.44 & 75.74 \ensuremath{\pm} 0.43 & 61.71 \ensuremath{\pm} 0.20 \\
& EPPC-OASIS & 86.63 \ensuremath{\pm} 0.41 & 78.62 \ensuremath{\pm} 0.43 & 84.32 \ensuremath{\pm} 0.08 & 75.43 \ensuremath{\pm} 0.06 & 61.49 \ensuremath{\pm} 0.23 \\
& EPPC-OASIS (weight avg.) & \bestresult{87.29} & \bestresult{80.34} & \bestresult{85.59} & \bestresult{77.13} & \bestresult{63.83} \\
\midrule
DeepSeek-R1-Distill-Llama-70B & SFT & 84.98 \ensuremath{\pm} 0.64 & 78.03 \ensuremath{\pm} 0.95 & 81.94 \ensuremath{\pm} 0.29 & 74.64 \ensuremath{\pm} 1.01 & 60.45 \ensuremath{\pm} 1.00 \\
& EPPC-OASIS & 84.04 \ensuremath{\pm} 0.70 & 75.80 \ensuremath{\pm} 0.91 & 81.09 \ensuremath{\pm} 0.21 & 72.38 \ensuremath{\pm} 0.28 & 58.66 \ensuremath{\pm} 0.86 \\
& EPPC-OASIS (weight avg.) & \bestresult{85.28} & \bestresult{78.62} & \bestresult{83.01} & \bestresult{75.31} & \bestresult{61.28} \\
\bottomrule
\end{tabular}
\end{table*}

\begin{table}[t]
\caption{Diagnostic component-wise best scores across inference stages. This table is an upper-bound analysis and is not used as the primary deployable result when different components come from different inference stages.}
\label{tab:component-wise-best}
\centering
\scriptsize
\setlength{\tabcolsep}{2pt}
\begin{tabular}{@{}>{\raggedright\arraybackslash}p{0.40\columnwidth}ccc@{}}
\toprule
\makecell{\textbf{Model}} & \makecell{\textbf{Best Code}\\\textbf{F1 (\%) \(\uparrow\)}} & \makecell{\textbf{Best Sub-code}\\\textbf{F1 (\%) \(\uparrow\)}} & \makecell{\textbf{Best Span}\\\textbf{F1 (\%) \(\uparrow\)}} \\
\midrule
Llama-3.2-1B-Instruct & 85.04 & 76.51 & 80.49 \\
Llama-3.2-3B-Instruct & 86.14 & 77.30 & 81.63 \\
Llama-3.1-8B-Instruct & 87.31 & 79.09 & 83.94 \\
Mistral-Small-24B-Instruct & \bestresult{87.49} & \bestresult{80.37} & \bestresult{84.64} \\
Qwen2.5-3B-Instruct & 85.43 & 77.71 & 81.97 \\
Qwen2.5-7B-Instruct & 85.97 & 79.29 & 83.50 \\
Qwen2.5-32B-Instruct & 87.29 & 80.34 & 85.59 \\
Llama-3.3-70B-Instruct & 86.92 & 78.96 & 83.94 \\
DeepSeek-R1-Distill-Llama-70B & 85.10 & 75.40 & 82.12 \\
\bottomrule
\end{tabular}
\end{table}

%
% \placeholdertable{Scaling results across model sizes.}{tab:scaling-results}{
% 1B & TODO: SFT vs EPPC-OASIS & README/final rerun & TODO \\
% 3B & TODO: SFT vs EPPC-OASIS & README/final rerun & TODO \\
% 8B & TODO: SFT vs EPPC-OASIS & README/final rerun & TODO \\
% 24B & TODO: SFT vs EPPC-OASIS & README/final rerun & TODO \\
% 70B & TODO: SFT vs EPPC-OASIS & README/final rerun & TODO \\
% }

\subsection{Ablation Studies}
\label{subsec:ablation-results}

\begin{figure*}[t]
\centering
\includegraphics[width=0.95\textwidth]{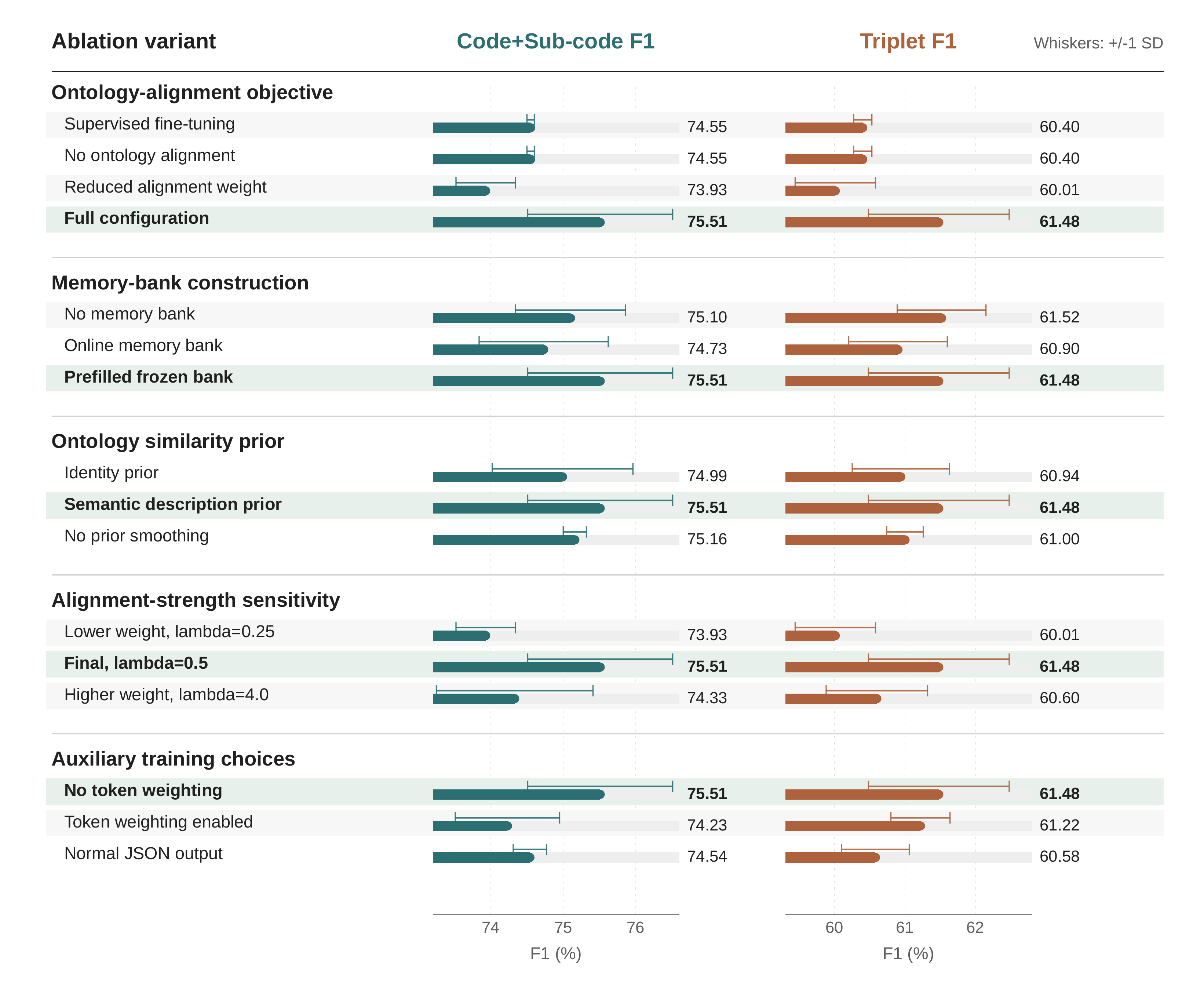}
\caption{Training ablation summary using Llama-3.1-8B-Instruct. Points show mean Code+Sub-code F1 and Triplet F1 with standard-deviation intervals across random seeds. Highlighted rows denote the final EPPC-OASIS configuration used in the main experiments.}
\label{fig:training-ablation-summary}
\end{figure*}

The main comparisons evaluate the final two-stage system, but they do not by themselves show which training and inference components are responsible for the observed gains. We therefore conducted ablation studies centered on Llama-3.1-8B-Instruct, using the same train-test split, output schema, parser, and evaluation metrics as in the primary comparison. Training ablations remove or modify one EPPC-OASIS design choice at a time, including the ontology-alignment objective, memory bank, ontology similarity prior, and alignment strength; these training ablations are summarized in Figure~\ref{fig:training-ablation-summary}. Inference ablations then compare greedy decoding, self-verification/refinement, self-consistency, hybrid label correction, and selector-based prediction. We focus on Code+Sub-code F1 and Triplet F1 because these metrics capture coherent EPPC label assignment and grounded structured extraction.

\subsubsection{Ontology-Alignment Objective}
\label{subsubsec:ablation-ontology-objective}

The first ablation evaluates whether the representation-level ontology objective provides benefit beyond ordinary supervised fine-tuning. We compare supervised fine-tuning with EPPC-OASIS variants in which the ontology-alignment term is removed, weakened, or applied in its final configuration while keeping the same base model, LoRA adaptation setup, decoding, parser, and evaluation procedure. A consistent drop after removing the ontology term would indicate that the method's benefit is not simply due to task-specific fine-tuning, but to explicitly shaping model representations according to the EPPC ontology.

\subsubsection{Memory-Bank Alignment}
\label{subsubsec:ablation-memory-bank}

The memory-bank ablation examines whether ontology alignment benefits from a broader set of comparison examples than those accumulated during ordinary training. This is important because EPPC labels are imbalanced and multi-label, so related examples may be absent from the early online bank even when they are present in the training set. We compare the Wasserstein trainer with no memory bank, with an online bank populated during training, and with the prefilled frozen full training-set bank used in the final configuration. Consistent with our design goal of maximizing hierarchical label understanding over pure span generation, removing the memory bank entirely produced a slightly higher Triplet F1 (61.52) but degraded the primary structural metric (Code+Sub-code F1 fell from 75.51 to 75.10). This indicates that the prefilled frozen bank is necessary to stabilize the alignment of the complex label hierarchy, even if it trades off marginal performance on exact token boundaries.

\subsubsection{Ontology Similarity Prior}
\label{subsubsec:ablation-ontology-prior}

The ontology-prior ablation tests whether EPPC-OASIS benefits from graded similarity among label descriptions beyond exact code/sub-code overlap. The identity-prior variant treats only exact shared labels as related, whereas the semantic-prior variant allows examples with related EPPC descriptions to exert partial alignment pressure. This comparison evaluates whether description-based smoothing improves learning in a label space where some communication behaviors are semantically adjacent but not identical.

\iffalse
\begin{table}[t]
\caption{Ablation of the ontology similarity prior.}
\centering
\scriptsize
\setlength{\tabcolsep}{3pt}
\begin{tabular}{@{}>{\raggedright\arraybackslash}p{0.54\columnwidth}cc@{}}
\toprule
\textbf{Variant} & \makecell{\textbf{Code+Sub-code}\\\textbf{F1 (\%) \(\uparrow\)}} & \makecell{\textbf{Triplet}\\\textbf{F1 (\%) \(\uparrow\)}} \\
\midrule
Identity prior & 74.99 \ensuremath{\pm} 0.97 & 60.94 \ensuremath{\pm} 0.69 \\
Semantic description prior & 74.72 \ensuremath{\pm} 0.33 & 61.00 \ensuremath{\pm} 0.25 \\
No prior smoothing & 75.16 \ensuremath{\pm} 0.16 & 61.00 \ensuremath{\pm} 0.26 \\
\bottomrule
\end{tabular}
\end{table}
\fi

\subsubsection{Alignment-Strength Sensitivity}
\label{subsubsec:ablation-alignment-strength}

We varied the ontology-loss weight and Sinkhorn regularization around the final configuration to examine the stability of the Wasserstein alignment objective. This analysis is intended to distinguish a robust ontology-alignment effect from a narrow hyperparameter optimum. The final ontology weight ($\lambda_{\mathrm{ont}}$) was selected based on validation subset performance prior to test-set evaluation. The experiment demonstrates that stable performance is maintained near this selected value, but structural coherence falls off if the alignment pressure is too weak (e.g., 0.25) or excessively dominates the supervised extraction loss (e.g., 4.0), underscoring that the representation-neighborhood alignment signal requires coarse calibration but is not dependent on highly precise tuning.

\iffalse
\begin{table}[t]
\caption{Sensitivity to ontology-alignment strength.}
\centering
\scriptsize
\setlength{\tabcolsep}{2.5pt}
\begin{tabular}{@{}>{\raggedright\arraybackslash}p{0.34\columnwidth}cccc@{}}
\toprule
\textbf{Variant} & \(\boldsymbol{\lambda_{\mathrm{ont}}}\) & \makecell{\textbf{Sinkhorn}\\\textbf{reg.}} & \makecell{\textbf{Code+Sub-code}\\\textbf{F1 (\%) \(\uparrow\)}} & \makecell{\textbf{Triplet}\\\textbf{F1 (\%) \(\uparrow\)}} \\
\midrule
Lower alignment weight & 0.25 & 0.01 & 73.93 \ensuremath{\pm} 0.41 & 60.01 \ensuremath{\pm} 0.57 \\
Final configuration & 0.5 & 0.01 & 75.51 \ensuremath{\pm} 1.00 & 61.48 \ensuremath{\pm} 1.00 \\
Higher alignment weight & 4.0 & 0.01 & 74.33 \ensuremath{\pm} 1.08 & 60.60 \ensuremath{\pm} 0.72 \\
Alternative regularization & N/A & N/A & N/A & N/A \\
\bottomrule
\end{tabular}
\end{table}
\fi

\subsubsection{Auxiliary Training Choices}
\label{subsubsec:ablation-auxiliary-training}

Finally, we evaluated auxiliary training choices that may affect performance but are not part of the core EPPC-OASIS contribution. These include token weighting and output-format variants used during model development. We report them separately from the main ablations to distinguish improvements attributable to ontology-aware training from improvements caused by general implementation choices or prompt/output formatting decisions.

\section{Discussion}
\label{sec:discussion}

\subsection{Principal Findings}
\label{subsec:principal-findings}

In this study, we evaluated a two-stage approach for structured extraction of patient-provider communication behaviors from secure messages. The first stage, EPPC-OASIS, incorporates EPPC ontology structure during fine-tuning so that examples with related codes and sub-codes are organized more coherently in representation space. The second stage applies structured inference refinement to verify, resample, merge, or select complete JSON predictions before scoring. 

The main finding was that this combination improved structural extraction over prompting-only baselines and standard supervised fine-tuning. Improvements were most evident on metrics that evaluate hierarchical label consistency, particularly Code F1, Sub-code F1, and joint Code+Sub-code F1. EPPC-OASIS achieved the highest structural label coherence among evaluated methods. While token-level preference methods like DPO were highly competitive for exact span localization and thus scored slightly higher on the strictly grounded Triplet F1 metric EPPC-OASIS best preserved the complex label hierarchy. This expected trade-off occurs because EPPC-OASIS optimizes the ontology representation space rather than explicitly optimizing token-level sequence generation, underscoring why structural label extraction and exact evidence grounding often require distinct methodological treatments.

The secondary analyses were consistent with this interpretation. Across model families and scales, the final deployable EPPC-OASIS pipelines produced consistent improvements, with the largest relative gains observed for smaller models (e.g., Llama-3.2-1B). Training ablations showed that removing ontology alignment, the memory bank, or the semantic description prior reduced performance on Code+Sub-code F1 and Triplet F1, indicating that the benefit was not fully explained by the shared fine-tuning setup. Inference ablations further showed that selector-style aggregation and CoT-SR contributed additional gains by correcting different failure modes after training. Diagnostic analyses showed high schema adherence without increasing formatting failures. A detailed error analysis of residual label and evidence-grounding failures is provided in Appendix~\ref{app:error-analysis}.

\subsection{Interpretation of Ontology-Aware Alignment}
\label{subsec:interpretation}

The results suggest that EPPC-OASIS is useful because it changes how training examples are organized during fine-tuning. Standard sequence-level fine-tuning rewards the model for reproducing each target annotation, but it does not directly encode relationships among EPPC codes, sub-codes, and semantically related communication behaviors. EPPC-OASIS addresses this limitation by aligning each example's representation-neighborhood distribution with an ontology-derived neighborhood distribution over training examples. This mechanism is particularly relevant for EPPC extraction because the label space is hierarchical, multi-label, and imbalanced.

The memory-bank component is central to this interpretation. In ordinary minibatches, especially when fine-tuning large models, related EPPC examples may be absent or too sparse to provide a stable alignment signal. By maintaining a broader set of ontology-labeled representations, the memory bank allows the Wasserstein alignment objective to compare learned neighborhoods against ontology-derived neighborhoods from the training distribution rather than the accidental composition of a single minibatch. This positions EPPC-OASIS as a structured training signal that uses the EPPC ontology to guide how examples should relate to one another during adaptation, rather than as a generic regularizer.

\subsection{Interpretation of Structured Inference Refinement}
\label{subsec:interpretation-inference}

The inference-refinement results highlight a second property of EPPC extraction: labels and evidence spans often fail differently. A greedy prediction can identify a plausible span while assigning a neighboring sub-code, whereas a sampled prediction can recover the correct Code/Sub-code pair but produce a less stable or less exact span. Self-consistency is therefore useful primarily as a label-stabilization mechanism, while verification and refinement target evidence support and schema validity. Hybrid methods exploit this asymmetry by combining stable labels from repeated sampling with spans from greedy or refined predictions when those spans remain better grounded in the source text.

This interpretation also explains why component-wise best scores must be separated from deployable pipeline results. The highest Code F1, Sub-code F1, and Span F1 for a model may come from different inference stages, but such a component-wise combination does not correspond to a single system that could be applied prospectively. For that reason, the main results report only complete prediction pipelines, while component-wise best values are treated as diagnostic upper bounds. This distinction is important for clinical NLP reporting because a deployable structured extractor must produce one coherent annotation set for each message, not separately optimized label and span components.

\subsection{Clinical and Informatics Implications}
\label{subsec:clinical-implications}

Scalable EPPC extraction has potential value for clinical communication research because secure messages contain information about patient questions, clinician guidance, logistical barriers, social needs, emotional concerns, and shared decision-making that is difficult to characterize manually at large scale. By converting message text into structured EPPC annotations, systems such as EPPC-OASIS can support retrospective analyses of communication patterns across patient groups, clinical contexts, and care trajectories. The inclusion of supporting evidence spans also improves auditability, allowing researchers to inspect the text underlying predicted communication behaviors rather than relying only on document-level labels.

The main implication of the proposed approach is therefore informatics enablement rather than immediate clinical automation. Compared with prompting or standard fine-tuning, ontology-aware training is intended to make structured EPPC extraction more reliable for downstream research workflows, particularly when labels are hierarchical, imbalanced, or semantically overlapping. Structured inference refinement adds an additional audit trail by making explicit which annotations were retained, corrected, merged, or dropped during post-training prediction. This may be useful for studying unmet informational, emotional, logistical, or social needs in patient-provider communication and for developing future tools for communication-quality measurement, cohort-level monitoring, or annotation support. 

For retrospective communication research, the achieved extraction performance demonstrates that EPPC-OASIS can substantially accelerate the analysis of large message corpora. While fully autonomous coding remains challenging due to the subtleties of clinical dialogue, a $\sim$77\% Code+Sub-code F1 indicates the model reliably organizes raw text into the correct hierarchical communication categories. In practice, this enables a highly efficient human-in-the-loop workflow: researchers can use the model to rapidly surface specific interactions such as complex shared decision-making or emerging Social Determinants of Health (SDOH), alongside the exact text spans that prompted the label.

Because false positives in sensitive categories could skew cohort-level estimates of patient need, the model's explicit extraction of grounded spans becomes a critical feature, allowing clinical reviewers to quickly audit and verify the evidence rather than reading messages from scratch. However, these applications require external validation before operational use. The present results do not establish that EPPC-OASIS improves clinical outcomes, message triage, treatment adherence, or care coordination; rather, they provide a methodological step toward more reliable analysis of patient-provider communication at scale.

\subsection{Relationship to Prior Work}
\label{subsec:relationship-prior-work}

This work builds on prior efforts to formalize EPPC extraction as a structured clinical NLP task. EPPCMinerBen established the benchmark setting and demonstrated that large language models can be evaluated on code, sub-code, and evidence-span extraction from patient-provider messages \cite{fodeh2026eppcminerben}. That work provided the task foundation for scalable EPPC mining, but its primary focus was benchmark construction and baseline evaluation rather than ontology-aware model adaptation. In contrast, the present study focuses on how the EPPC ontology can be used during training to improve structured extraction, particularly for joint code/sub-code prediction and grounded triplet extraction.

EPPC-OASIS is also related to STaR-DRO, which studied robustness for EPPC extraction under difficult or distributionally uneven groups \cite{fodeh2026stardro}. STaR-DRO and EPPC-OASIS address complementary aspects of reliability. STaR-DRO emphasizes robustness to group-level difficulty and distributional variation, whereas EPPC-OASIS uses the EPPC ontology to define a Wasserstein alignment target during fine-tuning. Both approaches are motivated by the same underlying challenge: EPPC labels are hierarchical, imbalanced, and semantically overlapping, so standard sequence-level supervision may not fully capture the structure of the task. More broadly, this work contributes to clinical LLM-based structured extraction by showing how domain ontologies can be incorporated into adaptation objectives rather than used only as output schemas or post hoc evaluation labels.

EPPC-OASIS also differs from preference-optimization approaches such as direct preference optimization and related ranking-based adaptation methods \cite{rafailov2024dpo,meng2024simpo}. Preference optimization can improve generation behavior by increasing the likelihood of preferred outputs relative to rejected alternatives, but it does not necessarily encode graded relationships among labels or examples unless those relationships are explicitly represented in the preference data. EPPC-OASIS instead uses the EPPC ontology to define neighborhood distributions over training examples and regularizes the model with a Sinkhorn optimal-transport objective during supervised adaptation \cite{cuturi2013sinkhorn}. The structured inference layer is related to self-consistency, self-refinement, and ensemble-style prediction methods \cite{wang2023selfconsistency,madaan2023selfrefine}, but it is specialized to EPPC extraction by treating labels and evidence spans as separable sources of error and by requiring the final output to remain a coherent Code/Sub-code/Span annotation set. The approach also complements broader clinical structured-extraction work in which ontologies or schemas are used to define valid outputs, normalize predictions, or evaluate extraction quality; here, the ontology is incorporated into both the adaptation objective and the inference-time interpretation of structured outputs.

\subsection{Limitations}
\label{subsec:limitations}

The results demonstrate the potential of ontology-aware fine-tuning for structured EPPC extraction, but several limitations should be considered when interpreting the findings. The study was conducted using secure patient-provider messages from a restricted clinical data environment. As a result, the distribution of message content, communication behaviors, and documentation practices may not reflect other institutions, specialties, or patient populations. External validation using independent datasets from other health systems, clinical specialties, and patient populations will therefore be necessary to determine whether the observed performance patterns generalize beyond the current corpus.

%The results demonstrate the potential of ontology-aware fine-tuning for structured EPPC extraction, but several limitations should be considered when interpreting the findings. The study was conducted using secure patient-provider messages from a restricted clinical data environment. As a result, the distribution of message content, communication behaviors, and documentation practices may not reflect other institutions, specialties, or patient populations. External validation will be needed to determine whether the observed performance patterns hold across different health systems and messaging contexts.

The EPPC ontology contains long-tailed labels, with some codes and sub-codes represented by relatively few examples. This imbalance affects both model training and evaluation: rare labels provide limited supervision, and their performance estimates are less stable than those for common communication behaviors. Although EPPC-OASIS was designed to use ontology structure to support learning across related labels, it does not eliminate the fundamental difficulty of sparse EPPC categories.

A further limitation is that the evaluation depends on the reference annotation schema and the consistency of manual labels. Secure messages can express several communication behaviors in a short span of text, and distinctions between related sub-codes may be subtle. Some apparent model errors may therefore reflect ambiguous or overlapping annotation boundaries rather than simple extraction failures. In addition, evidence spans were used to assess grounding, but span agreement should not be interpreted as a complete measure of clinical usefulness or communicative meaning.

Finally, the study was retrospective and did not evaluate clinical deployment. The strongest inference-refinement pipelines also require additional computation because they may generate multiple outputs per example, perform refinement passes, or merge predictions across seeds. This roughly $N$-fold increase in inference-time latency and compute is acceptable for the retrospective research workflows targeted in this study, but may necessitate optimization such as through distillation or early-exit criteria before high-throughput operational use in real-time clinical systems.

\subsection{Future Work}
\label{subsec:future-work}

A necessary next step is to evaluate EPPC extraction in clinical communication settings beyond the development corpus. The present study evaluates de-identified secure messages from one clinical data environment, whereas patient-provider messaging can vary substantially by institution, specialty, patient population, portal workflow, and local communication norms. Independently annotated corpora from additional health systems would therefore be needed to determine whether ontology-aware adaptation and structured inference refinement remain effective when the distribution of EPPC codes, sub-codes, and message styles changes. These evaluations should also move beyond aggregate performance by examining subgroup-level reliability, rare-category behavior, and site-specific error patterns, since an EPPC mining system intended for communication research must recover clinically meaningful behaviors consistently across heterogeneous patient and care-team contexts.

\section{Conclusion}
\label{sec:conclusion}

Reliable analysis of patient-provider messaging requires methods that can recover structured communication behaviors rather than only assign broad document-level labels. In this study, we evaluated EPPC-OASIS, an ontology-aware adaptation approach, together with structured inference refinement for automated EPPC extraction from de-identified secure messages. The proposed framework uses the EPPC code and sub-code hierarchy during model adaptation and then applies deployable inference procedures to improve the coherence of the final structured annotations. In the final evaluation, the best deployable pipeline achieved \textbf{77.13\%} Code+Sub-code F1 and \textbf{63.83\%} Triplet F1, corresponding to absolute gains of \textbf{+1.39} and \textbf{+2.12} F1 points over the strongest supervised fine-tuning baseline. These results support ontology-aware, structured extraction as a promising direction for scalable EPPC mining and retrospective clinical communication research, while emphasizing the need for external validation before operational use.

\section*{Declarations}

\subsection*{Code Availability}
The source code, training configurations, evaluation scripts, and non-PHI experimental artifacts used in this study are publicly available at: \url{https://anonymous.4open.science/r/EPPC-OASIS-5EA0/}.

\subsection*{Declaration of Competing Interest}
The authors declare that they have no known competing financial interests or personal relationships that could have appeared to influence the work reported in this paper.

\subsection*{Authors' Contributions}
Samah Fodeh: Conceptualization, Methodology, Resources, Supervision, Project administration, Writing -- original draft, Writing -- review and editing. Sreeraj Ramachandran: Methodology, Software, Formal analysis, Investigation, Writing -- original draft, Writing -- review and editing. Elyas Irankhah: Methodology, Software, Formal analysis, Investigation, Writing -- original draft, Writing -- review and editing. Muhammad Arif: Formal analysis, Investigation, Validation, Writing -- review and editing. Afshan Khan: Data curation, Writing -- review and editing. Ganesh Puthiaraju: Validation, Writing -- review and editing. Linhai Ma: Validation, Writing -- review and editing. Srivani Talakokkul: Data curation, Writing -- review and editing. Jordan Alpert: Validation, Writing -- review and editing. Sarah Schellhorn: Validation, Writing -- review and editing.

\subsection*{Acknowledgements}
We thank the clinical collaborators, annotators, and research staff who supported data curation, codebook refinement, and quality assurance for the EPPC Miner benchmark. Their domain expertise and sustained feedback were essential to the development of a high-fidelity structured annotation resource. This study used de-identified patient–provider secure-message data obtained from Yale New Haven Health (YNHH) and handled under institutional governance and applicable ethical oversight requirements. Data access, processing, and analysis were conducted in accordance with approved protocols and privacy safeguards.

\bibliographystyle{cas-model2-names}
\bibliography{references}

\clearpage
\appendix

\section{Appendix: Inference Details}
\label{app:prompts}

\subsection{Parser Logic}
Model outputs were normalized before scoring using a deterministic JSON recovery procedure. The parser removed surrounding Markdown code fences when present, attempted direct JSON deserialization, and then applied bracket-boundary recovery if direct parsing failed. Outputs that could not be decoded after recovery were treated as empty result lists. Parsed annotations were then validated against the EPPC code and sub-code inventory before metric computation.

\section{Extended Training Methodology}
\label{app:training-details}

\subsection{Hardware and Runtime}
Training and evaluation were conducted on institutional compute clusters with NVIDIA B200 and H200 accelerators. For 8B-parameter models, complete 3-epoch or 4-epoch fine-tuning runs typically required approximately 12--15 minutes per seed when using Unsloth-optimized kernels and 4-bit base-model quantization \cite{unsloth2024software}.

\subsection{LoRA Configurations}
Parameter-efficient fine-tuning used Low-Rank Adaptation (LoRA) across the primary transformer projection layers: \texttt{q\_proj}, \texttt{k\_proj}, \texttt{v\_proj}, \texttt{o\_proj}, \texttt{gate\_proj}, \texttt{up\_proj}, and \texttt{down\_proj}. Adapters used rank $r=32$, scaling factor $\alpha=64$, and dropout probability $0.05$. Optimization used 8-bit AdamW with cosine learning-rate decay.

\subsection{Preference-Pair Generation}
Preference pairs for the DPO and SimPO comparison baselines were generated from the EPPC gold training data \cite{rafailov2024dpo,meng2024simpo}. For each training example, the chosen response was the gold annotation set serialized in the required JSON format. The rejected response was produced by applying one of three synthetic corruptions with equal probability:
\begin{enumerate}
    \item \textbf{Annotation deletion:} a gold triplet was removed from the output.
    \item \textbf{Semantic substitution:} the sub-code was replaced with an incorrect sub-code from the same parent code hierarchy.
    \item \textbf{Span-boundary perturbation:} the evidence span was truncated or extended so that it no longer matched the gold span exactly.
\end{enumerate}

\subsection{Model-Specific Hyperparameters}

\begin{table}[H]
\caption{Model-specific training hyperparameters for final trainable configurations.}
\label{tab:training-hyperparameters}
\centering
\scriptsize
\setlength{\tabcolsep}{2pt}
\begin{tabular*}{\columnwidth}{@{\extracolsep{\fill}}lccc@{}}
\toprule
\makecell{\textbf{Model}} & \makecell{\textbf{Learning rate}} & \makecell{\(\boldsymbol{\lambda_{\mathrm{ont}}}\)} & \makecell{\textbf{Sinkhorn Reg.}} \\
\midrule
Llama-3.2-1B & \(8\times10^{-4}\) & 1.0 & 0.01 \\
Llama-3.2-3B & \(4\times10^{-4}\) & 0.5 & 0.01 \\
Qwen2.5-3B & \(4\times10^{-4}\) & 0.5 & 0.01 \\
Qwen2.5-7B & \(4\times10^{-4}\) & 0.25 & 0.01 \\
Llama-3.1-8B & \(4\times10^{-4}\) & 0.5 & 0.01 \\
Mistral-Small-24B & \(3\times10^{-4}\) & 0.5 & 0.01 \\
Qwen2.5-32B & \(4\times10^{-4}\) & 0.5 & 0.01 \\
Llama-3.3-70B & \(4\times10^{-4}\) & 0.5 & 0.01 \\
DeepSeek-R1-70B & \(4\times10^{-4}\) & 0.5 & 0.01 \\
\bottomrule
\end{tabular*}
\end{table}

\subsection{Ontology Prior Construction}
For the EPPC-OASIS semantic-prior ablation, the prior matrix $\mathbf{P}$ was constructed from textual EPPC label descriptions. Each label description concatenated the label name, its codebook definition, and definitions from adjacent hierarchy nodes where available. Descriptions were encoded with the base model's input embedding layer and mean-pooled over tokens to obtain one vector per code or sub-code. Pairwise prior similarity was then computed from the pooled label vectors.

\section{Error Analysis}
\label{app:error-analysis}

Aggregate F1 scores indicate whether a method improves extraction performance, but they do not show which kinds of mistakes are reduced. We therefore analyzed errors using categories aligned with the EPPC prediction structure. This analysis separates failures in broad communication-category recognition, fine-grained sub-code assignment, annotation recall, over-extraction, evidence grounding, and schema validity. Unless otherwise specified, error analyses compare supervised fine-tuning with greedy inference, EPPC-OASIS with greedy inference, and the final deployable EPPC-OASIS inference-refinement pipeline using the Llama-3.1-8B-Instruct reference model. Figure~\ref{fig:eppc_oasis_error_analysis} summarizes the residual sub-code confusions and comparative extraction error reductions achieved by EPPC-OASIS across evaluation levels.

\subsection{Error Taxonomy and Distribution}
\label{subsec:error-taxonomy}

We first categorized each unmatched or partially matched prediction according to its dominant failure mode. Code confusions were errors in the high-level EPPC communication category, whereas sub-code confusions captured cases in which the broad communicative function was recovered but the fine-grained behavior was incorrect. Missing annotations represented gold EPPC events with no corresponding prediction, while over-extraction represented predicted annotations without a matching gold event. Evidence-boundary errors were defined as cases in which the Code+Sub-code pair was correct but the predicted evidence span did not satisfy matching. Malformed JSON and invalid ontology labels were tracked separately because they reflect failures of schema adherence rather than ordinary classification or grounding errors.

\begin{table*}[t]
\caption{Detailed performance and error breakdown. Values are independent occurrence rates and do not sum to 100\%. Overall errors and adjacent-label confusions are percentages of test examples. Label validity errors are percentages of total predictions, rare-label omissions are percentages of rare gold instances, and grounding errors are computed only among correctly matched Code+Sub-code pairs.}
\label{tab:detailed-error-breakdown}
\centering
\scriptsize
\setlength{\tabcolsep}{4pt}
\begin{tabular}{@{}>{\raggedright\arraybackslash}p{0.34\textwidth}ccc@{}}
\toprule
\makecell{\textbf{Category /}\\\textbf{Error type}} & \makecell{\textbf{SFT}\\\textbf{greedy}} & \makecell{\textbf{EPPC-OASIS}\\\textbf{greedy}} & \makecell{\textbf{Difference /}\\\textbf{Rel. change}} \\
\midrule
\multicolumn{4}{@{}l}{\textit{Overall error rates (\% of test examples)}} \\
Code confusion & 47.31 \ensuremath{\pm} 0.31 & 45.13 \ensuremath{\pm} 0.36 & -2.18 pp \\
Sub-code confusion & 72.44 \ensuremath{\pm} 1.55 & 70.38 \ensuremath{\pm} 1.57 & -2.05 pp \\
Missing annotation & 67.56 \ensuremath{\pm} 1.01 & 64.10 \ensuremath{\pm} 0.48 & -3.46 pp \\
Over-extraction & 58.72 \ensuremath{\pm} 1.31 & 58.59 \ensuremath{\pm} 2.92 & -0.13 pp \\
Evidence-boundary error & 40.38 \ensuremath{\pm} 2.37 & 43.97 \ensuremath{\pm} 1.58 & +3.59 pp \\
Malformed JSON & 0.00 \ensuremath{\pm} 0.00 & 0.00 \ensuremath{\pm} 0.00 & +0.00 pp \\
Invalid ontology label & 0.55 \ensuremath{\pm} 0.30 & 0.92 \ensuremath{\pm} 0.47 & +0.36 pp \\
\addlinespace[3pt]
\midrule
\multicolumn{4}{@{}l}{\textit{Per-code performance (F1 \%)}} \\
InfoGive & 94.59 \ensuremath{\pm} 0.27 & 94.77 \ensuremath{\pm} 0.44 & +0.18 pp \\
InfoSeek & 76.13 \ensuremath{\pm} 2.35 & 76.87 \ensuremath{\pm} 3.24 & +0.74 pp \\
InfoGiveSDOH & 77.40 \ensuremath{\pm} 0.44 & 81.26 \ensuremath{\pm} 1.07 & +3.86 pp \\
Socioemotional/Empathy & 76.14 \ensuremath{\pm} 0.36 & 77.48 \ensuremath{\pm} 0.80 & +1.34 pp \\
PartnershipPatient & 92.81 \ensuremath{\pm} 0.69 & 92.27 \ensuremath{\pm} 0.65 & -0.54 pp \\
PartnershipProvider & 91.53 \ensuremath{\pm} 0.05 & 93.25 \ensuremath{\pm} 0.77 & +1.72 pp \\
SharedDecisionPatient & 41.72 \ensuremath{\pm} 10.50 & 46.87 \ensuremath{\pm} 4.95 & +5.15 pp \\
SharedDecisionProvider & 63.40 \ensuremath{\pm} 6.64 & 60.64 \ensuremath{\pm} 5.92 & -2.76 pp \\
\addlinespace[3pt]
\midrule
\multicolumn{4}{@{}l}{\textit{Ontology-consistency errors}} \\
Invalid code/sub-code pair & 0.62 \ensuremath{\pm} 0.34 & 1.11 \ensuremath{\pm} 0.37 & +0.49 pp \\
Parent/sub-code mismatch & 0.06 \ensuremath{\pm} 0.05 & 0.19 \ensuremath{\pm} 0.11 & +0.13 pp \\
Adjacent-label confusion & 28.85 \ensuremath{\pm} 1.13 & 27.44 \ensuremath{\pm} 0.48 & -1.41 pp \\
Rare-label omission & 70.83 \ensuremath{\pm} 9.00 & 69.44 \ensuremath{\pm} 1.96 & -1.39 pp \\
\addlinespace[3pt]
\midrule
\multicolumn{4}{@{}l}{\textit{Evidence-grounding errors (among correct labels)}} \\
Boundary drift & 14.69 \ensuremath{\pm} 0.51 & 16.20 \ensuremath{\pm} 0.52 & +1.51 pp \\
Wrong evidence phrase & 3.33 \ensuremath{\pm} 0.43 & 3.01 \ensuremath{\pm} 0.26 & -0.32 pp \\
No evidence span returned & 0.00 \ensuremath{\pm} 0.00 & 0.00 \ensuremath{\pm} 0.00 & +0.00 pp \\
\bottomrule
\end{tabular}
\end{table*}

Table~\ref{tab:detailed-error-breakdown} shows that ontology-aware training and structured inference refinement affected different parts of the error profile. EPPC-OASIS greedy inference reduced code confusions, sub-code confusions, missing annotations, adjacent-label confusions, and rare-label omissions relative to supervised fine-tuning. However, it also increased evidence-boundary errors and boundary drift among otherwise correct labels. This pattern suggests that ontology-aware training primarily improved label-structure behavior, while exact evidence localization remained a persistent limitation.

\subsection{Inference-Refinement Error Effects}
\label{subsec:inference-error-effects}

We separately examined errors introduced or corrected by inference refinement using the available label-corrected hybrid outputs. This analysis measured how often the hybrid output corrected labels while preserving an acceptable span, how often refinement corrected unsupported spans, and how often additional sampled annotations created over-extraction. These categories are important because a higher aggregate F1 can arise from qualitatively different behavior: improved label consensus, improved span grounding, or simply more aggressive prediction.

\begin{table}[t]
\caption{Error effects specific to structured inference refinement.}
\label{tab:inference-error-effects}
\centering
\scriptsize
\setlength{\tabcolsep}{3pt}
\begin{tabular}{@{}>{\raggedright\arraybackslash}p{0.44\columnwidth}>{\centering\arraybackslash}p{0.20\columnwidth}>{\raggedright\arraybackslash}p{0.28\columnwidth}@{}}
\toprule
\textbf{Inference effect} & \makecell{\textbf{Count}\\\textbf{/rate}} & \textbf{Interpretation} \\
\midrule
Label correction success & 9.62\% \ensuremath{\pm} 2.49 & Recovered at least one gold Code/Sub-code pair absent from greedy. \\
Span correction success & 33.97\% \ensuremath{\pm} 1.73 & Recovered acceptable grounding for a label found by greedy. \\
Added true-positive annotation & 8.46\% \ensuremath{\pm} 0.63 & Added at least one relaxed-triplet true positive. \\
Added false-positive annotation & 41.67\% \ensuremath{\pm} 1.31 & Added at least one unsupported relaxed-triplet false positive. \\
Dropped unsupported annotation & 35.64\% \ensuremath{\pm} 1.45 & Removed at least one greedy annotation not supported as a relaxed triplet. \\
Dropped valid annotation & 21.67\% \ensuremath{\pm} 1.48 & Removed at least one greedy relaxed-triplet true positive. \\
\bottomrule
\end{tabular}
\end{table}

\subsection{Label-Level Error Patterns}
\label{subsec:label-level-errors}

We next examined whether performance changes were concentrated in common labels or extended to less frequent EPPC behaviors. This analysis is important because the EPPC ontology is long-tailed: frequent categories provide many training examples, whereas rare but clinically meaningful behaviors may appear sparsely. Per-code and per-sub-code results therefore indicate whether EPPC-OASIS primarily improves dominant communication categories or whether ontology-aware alignment helps transfer information across related labels.

% \begin{table*}[t]
% \caption{Per-code performance for supervised fine-tuning and EPPC-OASIS.}
% \label{tab:per-code-errors}
% \centering
% \scriptsize
% \setlength{\tabcolsep}{5pt}
% \begin{tabular}{@{}>{\raggedright\arraybackslash}p{0.26\textwidth}cccc@{}}
% \toprule
% \textbf{EPPC code} & \textbf{Support} & \makecell{\textbf{SFT}\\\textbf{F1}} & \makecell{\textbf{EPPC-OASIS}\\\textbf{F1}} & \textbf{Difference} \\
% \midrule
% InfoGive & TODO & TODO & TODO & TODO \\
% InfoSeek & TODO & TODO & TODO & TODO \\
% InfoGiveSDOH & TODO & TODO & TODO & TODO \\
% InfoSeekSDOH & TODO & TODO & TODO & TODO \\
% EmotionalSupport & TODO & TODO & TODO & TODO \\
% CarePlanning & TODO & TODO & TODO & TODO \\
% Logistical & TODO & TODO & TODO & TODO \\
% Other & TODO & TODO & TODO & TODO \\
% \bottomrule
% \end{tabular}
% \end{table*}

As shown in Table~\ref{tab:detailed-error-breakdown}, EPPC-OASIS improved several high-level EPPC categories, with the largest gains for InfoGiveSDOH, SharedDecisionPatient, PartnershipProvider, and Socioemotional/Empathy. Performance was similar for InfoGive and InfoSeek, while SharedDecisionProvider decreased. This mixed pattern suggests that ontology-aware alignment helped several clinically meaningful label groups but did not eliminate instability for rare or semantically overlapping categories.

\subsection{Ontology Consistency}
\label{subsec:ontology-consistency-errors}

Because EPPC-OASIS explicitly uses the ontology during training, we separately evaluated errors related to schema consistency. These include invalid code/sub-code pairs, parent-code/sub-code mismatches, and confusions among semantically adjacent EPPC categories. Reductions in these errors would provide direct evidence that the method improves the internal consistency of structured annotations, beyond any improvement in individual label frequency or span matching.

% \begin{table}[t]
% \caption{Ontology-consistency errors for supervised fine-tuning and EPPC-OASIS.}
% \label{tab:ontology-consistency-errors}
% \centering
% \scriptsize
% \setlength{\tabcolsep}{3pt}
% \begin{tabular}{@{}>{\raggedright\arraybackslash}p{0.45\columnwidth}ccc@{}}
% \toprule
% \textbf{Error type} & \textbf{SFT} & \makecell{\textbf{EPPC}\\\textbf{OASIS}} & \makecell{\textbf{Relative}\\\textbf{change}} \\
% \midrule
% Invalid code/sub-code pair & TODO & TODO & TODO \\
% Parent/sub-code mismatch & TODO & TODO & TODO \\
% Adjacent-label confusion & TODO & TODO & TODO \\
% Rare-label omission & TODO & TODO & TODO \\
% \bottomrule
% \end{tabular}
% \end{table}

\subsection{Evidence Grounding Errors}
\label{subsec:grounding-errors}

Finally, we separated label correctness from evidence grounding. A prediction can recover the correct Code+Sub-code pair but still fail Triplet matching if the evidence span does not overlap sufficiently with the gold span. These cases are important because they indicate that the model has learned the semantic EPPC behavior but not the exact textual grounding expected by the annotation. We distinguish boundary drift from selection of the wrong evidence phrase where supported by the error analysis script.

% \begin{table}[t]
% \caption{Evidence-grounding errors among predictions with correct Code+Sub-code labels.}
% \label{tab:grounding-errors}
% \centering
% \scriptsize
% \setlength{\tabcolsep}{3pt}
% \begin{tabular}{@{}>{\raggedright\arraybackslash}p{0.45\columnwidth}ccc@{}}
% \toprule
% \textbf{Grounding error type} & \textbf{SFT} & \makecell{\textbf{EPPC}\\\textbf{OASIS}} & \makecell{\textbf{Relative}\\\textbf{change}} \\
% \midrule
% Boundary drift & TODO & TODO & TODO \\
% Wrong evidence phrase & TODO & TODO & TODO \\
% No evidence span returned & TODO & TODO & TODO \\
% \bottomrule
% \end{tabular}
% \end{table}

Grounding errors clarify how much of the remaining performance gap comes from evidence localization rather than label assignment. EPPC-OASIS slightly reduced wrong-evidence-phrase errors but increased boundary-drift errors among predictions with correct labels. These findings indicate that the ontology-aware objective improved parts of the label structure, but remaining errors are still substantially shaped by evidence-span localization.

\section{Model Diagnostics}
\label{app:diagnostics}

Model diagnostics were used to examine whether the training signals introduced by EPPC-OASIS behaved consistently with the method's intended mechanism. These analyses were not treated as additional task-performance endpoints; instead, they were used to interpret the relationship between ontology-aware alignment and the observed extraction results.

\subsection{Schema-Adherence Diagnostics}
\label{subsec:schema-diagnostics}

Schema-adherence diagnostics showed similar output validity for EPPC-OASIS and supervised fine-tuning. Both methods produced no unrecoverable JSON failures or invalid code/sub-code pairs under the final parser. EPPC-OASIS had a slightly higher invalid-label rate and a similar empty-output rate, indicating that the main performance differences were not driven by broad formatting failures. Table~\ref{tab:schema-diagnostics} summarizes these diagnostics.

\begin{table}[t]
\caption{Schema-adherence diagnostics for supervised fine-tuning and EPPC-OASIS.}
\label{tab:schema-diagnostics}
\centering
\scriptsize
\setlength{\tabcolsep}{3pt}
\begin{tabular}{@{}>{\raggedright\arraybackslash}p{0.46\columnwidth}cc@{}}
\toprule
\textbf{Diagnostic} & \makecell{\textbf{Supervised}\\\textbf{fine-tuning}} & \makecell{\textbf{EPPC}\\\textbf{OASIS}} \\
\midrule
Invalid JSON rate & 0.00\% \ensuremath{\pm} 0.00 & 0.00\% \ensuremath{\pm} 0.00 \\
Invalid label rate & 0.55\% \ensuremath{\pm} 0.30 & 0.92\% \ensuremath{\pm} 0.47 \\
Unrecoverable invalid code/sub-code pair rate & 0.00\% \ensuremath{\pm} 0.00 & 0.00\% \ensuremath{\pm} 0.00 \\
Empty-output rate & 1.28\% \ensuremath{\pm} 0.36 & 1.41\% \ensuremath{\pm} 0.36 \\
\bottomrule
\end{tabular}
\end{table}

\subsection{Train-Test Split Sensitivity}
\label{subsec:Train-Test Split Sensitivity}

To evaluate the sensitivity of the evaluation protocol to the choice of train-test partition, we compared three split configurations drawn from the same 867-example annotated pool. All experiments used Llama-3.2-3B-Instruct with LoRA-based supervised fine-tuning under the same hyperparameter configuration reported in Section\ref{subsec:implementation-details} (LoRA rank 32, scaling factor 64, dropout 0.05, 8-bit AdamW, cosine learning-rate schedule with $\text{lr} = 4 \times 10^{-4}$, effective batch size 16) and greedy decoding. Internal validation within the training portion used a fixed 90/10 ratio across all configurations. The split configurations differed only in the fraction of examples assigned to the external test partition. Table~\ref{tab:split_sensitivity} presents the resulting Code F1, Sub-code F1, Span F1, and parse rate for each split configuration.

The 70:30 split produced the highest extraction performance across all three component-level metrics (Code F1, Sub-code F1, and Span F1) and the highest output parse rate. The 80/20 configuration allocated more training examples but yielded a smaller test set (173 examples), which resulted in a substantially lower parse rate (87.9\% vs.\ 97.3\%) and reduced extraction performance on all metrics. The lower parse rate under the 80/20 split does not indicate overfitting to a particular test distribution, rather, it reflects the smaller training partition's reduced coverage of rare label combinations, which increases the likelihood of malformed outputs on infrequent EPPC patterns that the model encountered too few times during adaptation. The 70/15/15 three-way split reserved a separate development partition, further reducing the effective test size to 130 examples and likewise producing lower F1 scores. Because the EPPC label distribution is long-tailed with several clinically meaningful sub-codes appearing fewer than ten times in the corpus a larger test partition improves the reliability of per-label performance estimates and reduces the influence of individual difficult examples on aggregate metrics. 
Based on these results, the 70:30 stratified split (607 training examples, 260 test examples) was selected as the primary evaluation protocol for all experiments reported in the main text. This configuration provides sufficient training data for stable LoRA adaptation while preserving a test set large enough to support interpretable evaluation across the full EPPC label hierarchy.

\begin{table}[!h]
\caption{Effect of train-test split ratio on greedy SFT extraction performance using Llama-3.2-3B-Instruct.}
\label{tab:split_sensitivity}
\centering
\scriptsize
\setlength{\tabcolsep}{2pt}
\begin{tabular*}{\columnwidth}{@{\extracolsep{\fill}}lccccc@{}}
\toprule
\textbf{Split} & 
\makecell{\textbf{Train /}\\\textbf{Test}} & 
\makecell{\textbf{Code}\\\textbf{F1}} & 
\makecell{\textbf{Sub-code}\\\textbf{F1}} & 
\makecell{\textbf{Span}\\\textbf{F1}} & 
\makecell{\textbf{Parse}\\\textbf{rate}} \\
\midrule
70/30    & 607 / 260         & \textbf{80.38} & \textbf{75.17} & \textbf{77.88} & \makecell{253/260\\(97.3\%)} \\
80/20    & ${\sim}$694 / 173 & 75.34 & 64.09 & 67.04 & \makecell{152/173\\(87.9\%)} \\
70/15/15 & ${\sim}$607 / 130 & 74.20 & 67.05 & 66.13 & \makecell{122/130\\(93.8\%)} \\
\bottomrule
\end{tabular*}
\end{table}

\end{document}